\documentclass[journal]{IEEEtran}
\usepackage{amsmath,amsfonts}
\usepackage{algorithmic}
\usepackage{algorithm}
\usepackage{array}
\usepackage[caption=false,font=normalsize,labelfont=sf,textfont=sf]{subfig}
\usepackage{textcomp}
\usepackage{stfloats}
\usepackage{url}
\usepackage{verbatim}
\usepackage{graphicx}

\usepackage{graphicx}
\usepackage{subfig}  
\usepackage[table,xcdraw]{xcolor}
\usepackage{booktabs}
\usepackage{multirow}
\usepackage{amsmath, amsfonts}
\usepackage{latexsym}
\usepackage{times}   
\usepackage{soul}    
\usepackage{xspace}  
\usepackage{listings}
\usepackage[framemethod=default]{mdframed}
\usepackage{adjustbox}

\usepackage[most]{tcolorbox}
\usepackage{listings}
\newtcolorbox{exbox}[1]{
  enhanced,
  colback=gray!2,
  colframe=gray!55,
  boxrule=0.4pt,
  arc=2mm,
  left=1.2mm,right=1.2mm,top=0.9mm,bottom=0.9mm,
  title=\textbf{#1},
  fonttitle=\bfseries,
}

\usepackage[numbers,sort&compress]{natbib}
\usepackage{hyperref}
\usepackage{orcidlink} 
\hypersetup{
  hidelinks,
  pdftitle={Thought-Like-Pro: Enhancing Reasoning of Large Language Models through Self-Bootstrapped Prolog-based Chain-of-Thought},
  pdfauthor={Xihe Qiu; Yongxin Deng; Xiaoyu Tan; Zhen Fang; Ling Chen}
}

\newcommand{\method}{\textnormal{\textsc{Thought-Like-Pro}}\xspace}

\begin{document}

\title{\method: Enhancing Reasoning of Large Language Models through {Self-Bootstrapped} Prolog-based Chain-of-Thought\\[0.6em]
{\normalfont\footnotesize\parbox{0.96\textwidth}{\centering This article has been accepted for publication in IEEE Transactions on Cognitive and Developmental Systems (TCDS).\\[0.3em]
\copyright~2026 IEEE. Personal use of this material is permitted. Permission from IEEE must be obtained for all other uses, in any current or future media, including reprinting/republishing this material for advertising or promotional purposes, creating new collective works, for resale or redistribution to servers or lists, or reuse of any copyrighted component of this work in other works.}}}

\author{\thanks{\IEEEauthorrefmark{1}~Xihe Qiu and Yongxin Deng contributed equally to this work.}
\thanks{\IEEEauthorrefmark{2}~Corresponding author: Xihe Qiu (e-mail: \url{qiuxihe@u.nus.edu}; \url{qiuxihe1993@gmail.com}).}
Xihe Qiu\,\orcidlink{0000-0003-4024-925X}\IEEEauthorrefmark{1}\IEEEauthorrefmark{2},
Yongxin Deng\,\orcidlink{0009-0003-4352-1308}\IEEEauthorrefmark{1},
Xiaoyu Tan\,\orcidlink{0000-0003-3555-7143},\\
Zhen Fang\,\orcidlink{0000-0003-0602-6255},~\IEEEmembership{Member,~IEEE,}
and Ling Chen\,\orcidlink{0000-0002-6468-5729},~\IEEEmembership{Senior Member,~IEEE}
\thanks{This work was supported by the Shanghai Municipal Natural Science Foundation under Grant 23ZR1425400.}\thanks{Xihe Qiu and Yongxin Deng are with the School of Electronic and Electrical Engineering, Shanghai University of Engineering Science, Shanghai 201620, China.}\thanks{Xihe Qiu and Xiaoyu Tan are with the Department of Mechanical Engineering, National University of Singapore, Singapore 117575.}\thanks{Yongxin Deng, Zhen Fang, and Ling Chen are with the School of Computer Science, University of Technology Sydney, Sydney, NSW 2007, Australia.}}

\markboth{IEEE Transactions on Cognitive and Developmental Systems}{Qiu \MakeLowercase{\textit{et al.}}: Thought-Like-Pro}

\maketitle

\begin{abstract}

Large language models have demonstrated remarkable capabilities as general-purpose assistants, excelling in a wide range of reasoning tasks and supporting various aspects of daily web usage. This achievement represents a significant step toward achieving artificial general intelligence. Despite these advancements, the effectiveness of large language models often hinges on the specific prompting strategies employed, and there remains a lack of a robust framework to facilitate learning and generalization across diverse reasoning tasks. To address these challenges, we introduce a novel learning framework, \method. In this framework, we utilize imitation learning to imitate the Chain-of-\textbf{Thought} process which is verified and translated from reasoning trajectories generated by a symbolic \textbf{Pro}log logic engine. This framework proceeds in a {prompt-guided but self-bootstrapped} manner, that enables large language models to formulate rules and statements from given instructions and leverage the symbolic Prolog engine to derive results. Subsequently, large language models convert Prolog-derived successive reasoning trajectories into natural language chain-of-thought for imitation learning. The empirical findings indicate that our proposed approach greatly improves the reasoning capacity of large language models. By employing model averaging techniques, our method exhibits only a marginal decline in performance for distributional extrapolation tasks, showing robust generalization capabilities. We present a technical approach that integrates symbolic reasoning with language modeling, with the potential to support the development of large language models as cognitively inspired systems. The part of the dataset we used has been open-sourced.
\end{abstract}


\section{Introduction}

Recent advancements in large language models (LLMs) have solidified their reputation as versatile tools for a broad spectrum of tasks. This capability is the result of a structured training process that includes pre-training on extensive text corpora, supervised fine-tuning (SFT), and reinforcement learning from human feedback (RLHF) \citep{casper2023open}. In addition to general-purpose applications, LLMs have found significant usage in web-based environments, where they enhance semantic search, personalize content recommendations, streamline online customer support, and power tools for real-time content moderation and dynamic user engagement. These phases of training endow LLMs with vast knowledge, the ability to generate contextually relevant responses, and alignment with human values. A significant factor contributing to the view of LLMs as precursors to artificial general intelligence (AGI) is their scalable and explainable reasoning capabilities \citep{zhao2023survey}, which have shown promise in addressing complex real-world problems, including natural language logical reasoning and arithmetic reasoning. 

However, general logic reasoning tasks continue to present substantial challenges for LLMs. These challenges primarily emerge during the LLMs' reasoning processes \citep{ribeiro2023street, zhou2024survey}. At the semantic level, the generation of accurate multi-hop inferential pathways and the assurance of validity at each reasoning step remain significant obstacles \citep{creswell2023selectioninference, kazemi-etal-2023-lambada}. LLMs function in a self-regressive manner at the token level, processing information sequentially from left to right, one token at a time. This method can amplify errors: an inaccurate token prediction may negatively impact all subsequent tokens, leading to escalating inaccuracies and false reasoning outcomes. Interestingly, some studies \citep{ye2022the, saparov2023language, ribeiro2023street, deng2024promotingequalitylargelanguage} have observed instances where LLMs deliver correct answers despite erroneous internal reasoning processes. \textbf{This phenomenon highlights a misalignment between the reasoning processes and outcomes of LLMs, suggesting a weak causal relationship and underdeveloped implicit reasoning connections within the LLMs.}

To address the challenges associated with generating reasoning trajectories in LLMs, several strategies have been developed to enhance cognitive alignment and logical coherence. The Chain-of-Thought (CoT) approach \citep{wei2022chain, kojima2022large, zhang2022automatic}, for instance, enhances LLMs' reasoning abilities by crafting instructions and few-shot examples that promote sequential reasoning. Building on the CoT framework, other methodologies like Tree-of-Thoughts \citep{yao2024tree, long2023large} and Graph-of-Thoughts \citep{besta2024graph, zhang2024cumulative} have been introduced. These methods aim to formalize the reasoning architecture of LLMs through in-context learning (ICL) \citep{dong2022survey, workrethinking} and guide model output. While these prompt-based techniques have proven effective in specific tasks, they require precise, task-specific modifications and may not be broadly applicable across different domains \citep{creswell2023selectioninference, kazemi-etal-2023-lambada}. Alternative approaches have sought to augment the reasoning capabilities of LLMs by conceptualizing them as agents that can employ various logical reasoning tools \citep{lyu2023faithful, pan2023logic}. This strategy simplifies complex reasoning tasks into more manageable components such as tool utilization and question definition. One particularly effective method involves the use of symbolic logic engines, such as Prolog \citep{bowen1979prolog}, to conduct symbolic inference, thereby ensuring the correctness of the reasoning trajectory. However, this approach does not inherently enhance the LLM's reasoning capabilities and fails to provide a generalized improvement in the model's ability to follow general instructions across diverse tasks.

\begin{figure*}[ht]
  \centering
  \includegraphics[width=0.95\textwidth]{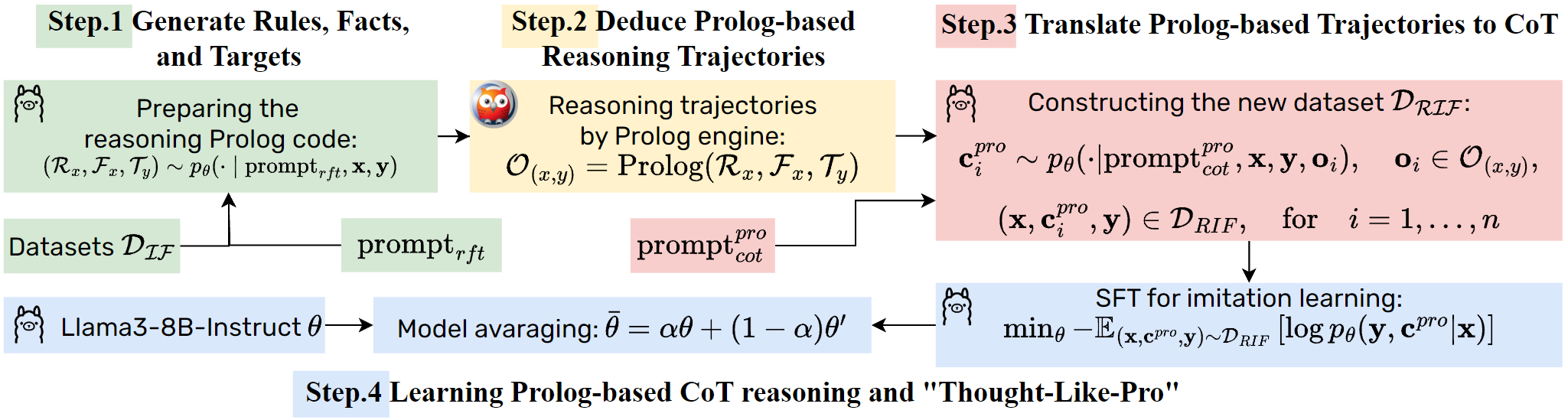}

  \caption{Workflow overview of our \method.}
  \label{fig:main}

\end{figure*}

In this paper, we introduce \method (TLP), a novel framework designed to enhance the logical reasoning capabilities of LLMs by imitating the verified reasoning processes of the Prolog logic engine. We begin by utilizing open-source chat LLMs to generate rules, facts, and queries {(more details in Appendix~\ref{app:prolog_rep})} following specific instructions and demonstrations. Subsequently, we employ the Prolog engine to perform logical reasoning, verifying the outcomes against established truths. Only the verified reasoning processes are selected, which we then translate into CoT-like natural language reasoning trajectories. These trajectories serve as the basis for supervised fine-tuning (SFT), aimed at teaching LLMs to imitate a \textbf{strictly logical reasoning process}. To address the issue of catastrophic forgetting during domain-specific fine-tuning, we implement a model averaging (MA) technique. Empirical experiments confirm that the \method framework significantly improves the reasoning capabilities of LLMs and extends their generalization to Out-of-Distribution (OOD) reasoning benchmarks. The contributions of our study are detailed as follows:
\begin{itemize}
\item We propose \method framework designed to enhance the general reasoning capabilities of LLMs in a self-driven way. This framework is straightforward to implement and replicate.
\item Our results show that performing imitation learning on \textbf{strictly logical and diverse reasoning trajectories for one single instruction} leads to improved and consistent reasoning performance.
\item The framework is \textbf{simple, straightforward, and highly effective.} It demonstrates robust OOD reasoning performance across multiple logic tasks and general benchmarks.
\item We evaluate our proposed \method on financial reasoning tasks within our company, demonstrating significant improvements over the baseline method, highlighting its potential applicability across various industrial domains.
\end{itemize}

\section{Related Work}
\subsection{Enhanced LLM Inference Capabilities}
The development of intelligent systems with reasoning capabilities has long been a central objective in the field of artificial intelligence \citep{wos1992automated, fagin2004reasoning}. Recent advances in LLMs, particularly the recognition of their emergent properties and in-context learning abilities, have opened new avenues for machine reasoning \citep{bubeck2023sparks, wei2022chain}. Several studies \citep{wei2022chain, drori2022neural} have employed CoT prompting and related strategies to guide LLMs in producing step-by-step solutions for mathematical and logical reasoning tasks. Building on these findings, recent research \citep{wang2023selfconsistency, fu2022complexity, zelikman2022star} has explored practices such as sampling multiple solutions and selecting the optimal response based on self-consistency or complexity criteria. For instance, Marco-O1 \citep{zhao2024marco} has utilized techniques including CoT fine-tuning, Monte Carlo Tree Search (MCTS), reflection mechanisms, and innovative reasoning strategies to extend the reasoning capabilities of LLMs in addressing complex real-world challenges. Nonetheless, much of this work still requires human intervention, prompting some researchers to investigate automated prompt generation methods, such as AutoPrompt \citep{shin2020autoprompt}, prefix-tuning \citep{li2021prefix}, parameter-efficient prompting \citep{lester2021power}, and automatic prompt engineering \citep{zhou2022large}.

In addition, a number of studies have focused on the interpretability of LLM reasoning. The chain-of-thought heuristic suggests that generating intermediate reasoning steps can benefit LLMs, thereby sparking interest in approaches that balance robust reasoning with interpretability. Some works have applied tree \citep{yao2024tree, long2023large} and graph structures \citep{besta2024graph, zhang2024cumulative}, demonstrating improved performance—particularly in complex reasoning tasks where the underlying processes are more intricate than a simple linear progression. Other recent research has taken a less structured approach \citep{tafjord2022entailer, creswell2023selectioninference, kazemi-etal-2023-lambada}; for example, Selection-Inference \citep{creswell2023selectioninference} proposed a framework that alternates between selection and inference, treating pre-trained LLMs as general processing units to generate a series of interpretable causal reasoning steps that ultimately yield the final answer.

Despite the fact that these reasoning proofs align with gold standard protocols, their reliance on LLM-generated outputs poses challenges in ensuring causal coherence and reliability—challenges that stem from the inherent tendency of LLMs to produce hallucinated content \citep{zhang2023hallucination, varshney2023stitch, lee2022factuality}. In contrast, our method employs a symbolic logic engine for symbolic reasoning, thereby producing reasoning steps that are logically irrefutable, with the engine itself verifying the correctness of the reasoning trajectory.

\subsection{Neuro-symbolic Reasoning}
In contrast to approaches that rely solely on LLMs to generate reasoning trajectories, neuro-symbolic systems strive to leverage the complementary strengths of neural networks and symbolic reasoning \citep{hudson2019learning, gupta2019neural, cobbe2021training}. These systems use neural networks for pattern recognition and learning from unstructured data while employing symbolic systems for rule-based reasoning and knowledge representation. Recent developments have showcased methods that integrate knowledge from unstructured human language via LLMs by converting natural language into symbolic representations \citep{lyu2023faithful, pan2023logic, yang2023neuro}. For example, Causal and Reliable Reasoning (CARING) \citep{yang2023neuro} combines a customized meta-interpreter with a symbolic solver to elucidate the underlying reasoning proofs. Although such methods mitigate errors in reasoning trajectories caused by LLM hallucinations, they fundamentally rely on auxiliary components rather than enhancing the intrinsic reasoning capabilities of LLMs. In contrast, our approach centers on using a deterministic neuro-symbolic system to correct and refine the diverse reasoning trajectories produced by LLMs, thereby enhancing the model’s performance on specific tasks.

\section{Methods}
\label{sec:methods}

\subsection{Preliminary: Supervised Fine-tuning, Chain-of-Thought, and Prolog}

Consider a LLM, denoted by $p_\theta$, which is parameterized by $\theta$. Given an instruction dataset $\mathcal{D}_{IF}$ consisting of question and answer pairs $(\mathbf{x}, \mathbf{y})$, a highly effective method for aligning $\theta$ with the target output $\mathbf{y}$ given $\mathbf{x}$ is to minimize the negative log-likelihood of $\mathbf{y}$ conditioned on $\mathbf{x}$ \citep{gunel2020supervised, dong2023abilities}. This can be formulated autoregressively as:
\begin{equation}
\label{equ:sft}
    \min_\theta -\mathbb{E}_{(\mathbf{x}, \mathbf{y}) \sim \mathcal{D}_{IF}} \left[\sum_{i=1}^k \log p_\theta(y_i|\mathbf{x})\right],
\end{equation}
where $k$ is the number of tokens in $\mathbf{y}$.

To enhance reasoning performance on complex logical and algorithmic tasks, one effective approach involves using a specialized prompt, \({\rm prompt}_{\text{CoT}}\), to initiate a step-by-step reasoning trajectory prior to generating the final output: \((\mathbf{y}, \mathbf{c}) \sim p_\theta(\cdot|\mathbf{x}, {\rm prompt}_{\text{CoT}})\). Here, \(\mathbf{c}\) represents a CoT reasoning process \citep{wei2022chain, kojima2022large, zhang2022automatic}. Typically, this process decomposes complex, multi-step reasoning tasks into simpler, intermediate steps, thereby directing the LLM toward the correct answer. Nevertheless, several studies \citep{ye2022the, saparov2023language, ribeiro2023street} have indicated that \(\mathbf{c}\) may sometimes exhibit disorganized patterns that do not adhere strictly to logical reasoning processes. This inconsistency can adversely affect the LLM's reasoning performance across diverse tasks.

Prolog is a high-level symbolic language system designed for rule-based reasoning \citep{7780381, neelakantan2017learning}, fundamentally operating within a subset of predicate logic known as Horn Clause logic \citep{hudson2019learning, bowen1979prolog}. It is renowned for its declarative programming paradigm, which contrasts with the imperative approach by emphasizing relations over the sequence of actions. In Prolog, the logic of computation is articulated through relations, denoted as facts $\mathcal{F}$ and rules $\mathcal{R}$. Typically, Prolog utilizes a depth-first search strategy \citep{wielemaker2012swi} to trace a feasible logical inference path to the desired conclusion, yet it is also equipped to delineate all possible logical inference paths leading to the target results $\mathcal{T}$. The entire logical inference process within the Prolog engine, denoted as ${\rm Prolog}$, can be represented by the following equation:
\begin{equation}
\label{equ:prolog}
    \mathcal{O} = {\rm Prolog}(\mathcal{R}, \mathcal{F}, \mathcal{T}), \{o_1, o_2,...,o_n\} \in \mathcal{O},
\end{equation}
where $\mathcal{O}$ encompasses $n$ valid logical inference trajectories toward the target $\mathcal{T}$.

\subsection{\method}
\label{Thought-Like-Pro}
Here, we introduce \method, a novel framework designed to enhance the general logical reasoning capabilities of LLMs across various logical tasks. As suggested by the name, our framework enables LLMs to imitate strictly logical reasoning trajectories that are generated and validated by the Prolog engine. As shown in Fig.~\ref{fig:main}, the architecture of \method is organized in a self-driven manner, eliminating the requirement for auxiliary services from other advanced LLMs, such as GPT-4 \citep{achiam2023gpt} and Claude 2 \citep{wu2023comparative}. It solely relies on the utilization of efficient, open-source LLMs (i.e., Llama3-8B-Instruct \citep{dubey2024llama}), fostering accessibility and ease of replication.

To employ the Prolog engine $\mathrm{Prolog}$ for logical reasoning, we initiate the process by constructing a few-shot demonstration $\mathrm{prompt}_{rft}$ (Appendix~\ref{Prompt demo}). This is designed to guide the model $p_\theta$ in generating relevant rules $\mathcal{R}_{x}$, facts $\mathcal{F}_{x}$, and deriving the reasoning target $\mathcal{T}_{y}$:
\begin{equation}
\begin{aligned}
    (\mathcal{R}_{x}, \mathcal{F}_{x}, \mathcal{T}_{y}) & \sim p_\theta(\cdot \mid \mathrm{prompt}_{rft}, \mathbf{x}, \mathbf{y}).
\end{aligned}
\label{equ:trans}
\end{equation} Subsequently, the Prolog engine is employed to deduce the set of reasoning trajectories $\mathcal{O}_{(x,y)}$, which encompasses all logical paths leading to the target $\mathcal{T}_{y}$. This process is formalized as:
\begin{equation}
    \mathcal{O}_{(x,y)} = \mathrm{Prolog}(\mathcal{R}_{x}, \mathcal{F}_{x}, \mathcal{T}_{y}).
\end{equation} During this phase, we selectively utilize only those reasoning trajectories that conclusively reach the target $\mathcal{T}_{y}$, excluding any data that result from erroneous transformations or generation failures as delineated in Equation~\ref{equ:trans}. To ensure the acquisition of high-quality and interpretable reasoning trajectories via Prolog, we implement a meta-solver developed by \citep{yang2023neuro}, which retrieves all feasible reasoning paths.

After acquiring the set $\mathcal{O}_{(x,y)}$, we design a novel few-shot prompt, ${\rm prompt}_{cot}^{pro}$ (Appendix~\ref{Prompt demo}), to translate existing reasoning trajectories into CoT-like reasoning processes expressed in natural language. This transformation aids in constructing the new dataset $\mathcal{D}_{RIF}$:
\begin{equation}
\begin{aligned}
    &\mathbf{c}^{pro}_i \sim p_\theta(\cdot|{\rm prompt}_{cot}^{pro}, \mathbf{x}, \mathbf{y}, \mathbf{o}_i), \mathbf{o}_i \in \mathcal{O}_{(x,y)}, \\
    &(\mathbf{x}, \mathbf{c}^{pro}_i, \mathbf{y}) \in \mathcal{D}_{RIF}, \quad \text{for} \quad i =1,...,n. 
\end{aligned}
\end{equation} It is important to note that this new dataset, $\mathcal{D}_{RIF}$, encompasses all $n$ available reasoning trajectories for each instruction $\mathbf{x}$. Subsequently, we optimize the model using SFT as outlined in Equation~\ref{equ:sft}, aligning both the CoT-like reasoning trajectories and the target outputs conditioned on $\mathbf{x}$ autoregressively:
\begin{equation}
\label{equ:sft_pro}
\min_\theta -\mathbb{E}_{(\mathbf{x}, \mathbf{c}^{pro}, \mathbf{y}) \sim \mathcal{D}_{RIF}}\left[\log  p_\theta(\mathbf{y}, \mathbf{c}^{pro}|\mathbf{x})\right],
\end{equation}
resulting in a trained model parameterized by $\theta^\prime$. This entire learning process parallels imitation learning algorithms \citep{hussein2017imitation, osa2018algorithmic} that select ``optimal'' trajectories (here verified by the Prolog engine) and implement behavior cloning to directly imitate these trajectories. After training according to Equation~\ref{equ:sft_pro}, the model $p_{\theta^\prime}$ is tailored to perform domain-specific tasks present in $\mathcal{D}_{RIF}$. However, if the task distribution of $\mathcal{D}_{RIF}$ significantly diverges from that of $\mathcal{D}_{IF}$, catastrophic forgetting (CF) may occur \citep{chen2023challenges, zhai2023investigating, shi2024continual}, potentially compromising the model’s generality across other reasoning and general tasks. To address this, we employ a model averaging technique \citep{lin2023speciality} to derive a new model $p_{\bar{\theta}}$ parameterized by $\bar{\theta}$: $\bar{\theta} = \alpha \theta + (1-\alpha) \theta^\prime$, where $\alpha$ is a hyper-parameter that balances specialization and generalization.

\begin{table*}[t]
\caption{\textbf{Performance on Prolog-Tractable Reasoning Tasks.}}
\vspace{2mm}
\centering
\label{tab:main_tractable}
\setlength{\tabcolsep}{4.2pt}
\renewcommand{\arraystretch}{1.18}
\resizebox{\linewidth}{!}{
\begin{tabular}{lcccccccccc}
\toprule
\multirow{3}{*}{\textbf{Method}} & \multicolumn{10}{c}{\textbf{Acc $\uparrow$}} \\
\cmidrule(lr){2-11}
& \multicolumn{2}{c}{\textbf{GSM8K}} & \multicolumn{2}{c}{\textbf{ProofWriter}} & \multicolumn{2}{c}{\textbf{PrOntoQA}} & \multicolumn{2}{c}{\textbf{QASC}} & \multicolumn{2}{c}{\textbf{Average}} \\
\cmidrule(lr){2-3}\cmidrule(lr){4-5}\cmidrule(lr){6-7}\cmidrule(lr){8-9}\cmidrule(lr){10-11}
& \textbf{Llama3-8B} & \textbf{Qwen3-8B} & \textbf{Llama3-8B} & \textbf{Qwen3-8B} & \textbf{Llama3-8B} & \textbf{Qwen3-8B} & \textbf{Llama3-8B} & \textbf{Qwen3-8B} & \textbf{Llama3-8B} & \textbf{Qwen3-8B} \\
\midrule
\rowcolor{gray!15}
\textbf{Prompt} & \multicolumn{2}{c}{8-shot, CoT} & \multicolumn{2}{c}{3-shot, CoT} & \multicolumn{2}{c}{2-shot} & \multicolumn{2}{c}{0-shot} & \multicolumn{2}{c}{--} \\
\midrule

\rowcolor{gray!30} \multicolumn{11}{l}{\textbf{Training-free baselines}} \\
Base model & 79.60 & 89.81 & 53.70 & 57.95 & 53.60 & 62.50 & 95.60 & 94.20 & 70.63 & 76.12 \\
Self-Consistency & 80.10 & 90.25 & 61.20 & 63.70 & 87.60 & 86.40 & 96.00 & 95.80 & 81.23 & 84.04 \\
CoT-decoding$^{\dagger}$ & 79.30 & 85.70 & 52.70 & 55.40 & 59.50 & 60.70 & 96.30 & 94.50 & 71.95 & 74.08 \\

\rowcolor{gray!30} \multicolumn{11}{l}{\textbf{Training-based baselines}} \\
Ground-truth training$^{\ddagger}$ & 80.50 & 90.20 & 64.40 & 64.41 & 89.90 & 93.20 & \textcolor{gray}{N/A} & \textcolor{gray}{N/A} & 78.27 & 82.60 \\
STaR & 79.85 & 89.75 & 63.14 & 58.05 & 83.10 & 84.90 & 95.60 & 94.50 & 80.42 & 81.80 \\
CoT knowledge distillation & 79.89 & 90.41 & 63.20 & 58.17 & 85.20 & 87.60 & 95.90 & 96.90 & 81.05 & 83.27 \\
\rowcolor{blue!10} TLP (MA+single) & 80.70 & 90.60 & 72.13 & 73.33 & 90.40 & 94.12 & 96.00 & 96.20 & 84.81 & 88.56 \\
\rowcolor{blue!10} TLP (MA+multiple) & \underline{85.32} & 93.60 & \underline{97.25} & \underline{98.75} & 94.30 & \underline{97.10} & \underline{96.50} & \underline{97.10} & \underline{93.34} & \underline{96.64} \\
\rowcolor{blue!10} TLP (single) & 83.40 & \underline{95.02} & 73.91 & 79.30 & \underline{95.30} & 95.80 & 96.10 & 97.00 & 87.18 & 91.80 \\
\rowcolor{blue!10} TLP (multiple) & \textbf{87.81} & \textbf{95.54} & \textbf{98.19} & \textbf{99.75} & \textbf{100.00} & \textbf{100.00} & \textbf{96.70} & \textbf{98.30} & \textbf{95.68} & \textbf{98.40} \\

\midrule[1.5pt]
\rowcolor{gray!30} \multicolumn{11}{l}{\textbf{LLMs as Agents}} \\
CARING & \textcolor[HTML]{9B9B9B}{98.10} & \textcolor[HTML]{9B9B9B}{99.40} & \textcolor[HTML]{9B9B9B}{100.00} & \textcolor[HTML]{9B9B9B}{100.00} & \textcolor[HTML]{9B9B9B}{100.00} & \textcolor[HTML]{9B9B9B}{100.00} & \textcolor[HTML]{9B9B9B}{97.60} & \textcolor[HTML]{9B9B9B}{99.60} & \textcolor[HTML]{9B9B9B}{98.93} & \textcolor[HTML]{9B9B9B}{99.75} \\

\bottomrule
\end{tabular}
}

\begin{minipage}{0.985\textwidth}
\vspace{1mm}
\footnotesize
{\textbf{Notes.} `Acc' denotes accuracy; N/A denotes not applicable; STaR denotes Self-Taught Reasoner. GSM8K: Grade School Math 8K; PrOntoQA: Proof and Ontology-Generated Question-Answering; QASC: Question Answering via Sentence Composition. We report accuracy on four tractable reasoning benchmarks with two LLM backbones. Best results are \textbf{bolded} and second-best are \underline{underlined} (ranked among non-agent baselines) \emph{per column}. CARING results are shown in grey and not ranked, as they represent an approximate upper bound of pure symbolic reasoning on tractable tasks.
$^{\dagger}$ CoT-decoding elicits reasoning via decoding (no explicit CoT exemplars in the prompt).
$^{\ddagger}$ QASC does not contain ground-truth reasoning paths, hence supervised training on QASC is not applicable (N/A).
Averages are computed over available datasets (excluding N/A).}
\end{minipage}
\vspace{-2mm}
\end{table*}

\section{Experiments}

In our experiments, we evaluated the \method{}
framework's effectiveness in enhancing the reasoning capabilities
of LLMs. Before presenting the research questions, we clarify how
the term \emph{reasoning} is used in this paper, since multiple
interpretations exist in the literature.
{

Following the notation in Section~\ref{sec:methods}, let $p_\theta$
denote the LLM under evaluation, and $(x, y)$ a question-answer
pair drawn from a reasoning task distribution $\mathcal{D}$. We
say that $p_\theta$ exhibits \emph{intrinsic reasoning} on
$\mathcal{D}$ if it produces an answer $\hat{y}$ that semantically
aligns with $y$ by conditioning only on $x$ (optionally with a CoT
prompt $\mathrm{prompt}_{\mathrm{CoT}}$), without invoking the
symbolic solver $\mathrm{Prolog}(\cdot)$ defined in
Eq.~(\ref{equ:prolog}) or any external symbolic engine at inference
time. The corresponding accuracy is
\begin{equation}
\label{eq:intrinsic-acc}
    \mathrm{Acc}(p_\theta;\mathcal{D})
    =
    \mathbb{E}_{(x,y)\sim\mathcal{D}}
    \!\left[\mathbb{I}\!\left(\hat{y}\text{ semantically aligns with }y\right)\right],
\end{equation}
and we say that $p_\theta$ exhibits \emph{stronger} intrinsic
reasoning than $p_{\theta'}$ on $\mathcal{D}$ when
$\mathrm{Acc}(p_\theta;\mathcal{D}) > \mathrm{Acc}(p_{\theta'};\mathcal{D})$
under this tool-free inference mode.

This definition is behavioral and operational. It treats intrinsic
reasoning as a property of end-task accuracy under tool-free
inference, and makes no claim about a mechanistically distinct
``reasoning faculty'' inside the model. Agentic baselines such as
CARING~\cite{yang2023neuro}, which invoke
$\mathrm{Prolog}(\mathcal{R}_x, \mathcal{F}_x, \mathcal{T}_y)$ at
test time, fall outside this scope and are reported separately for
reference. We discuss the limits of this behavioral view in
Section~\ref{sec:limitations}.
}

Our investigation was guided by three key research questions (RQs):

\begin{itemize}
    \item \textbf{RQ1:} How effectively does \method enhance LLM reasoning capabilities on structured reasoning tasks with explicit logical structures?
    \item \textbf{RQ2:} Can \method improve the LLM's intrinsic ability to handle reasoning that requires implicit knowledge or deals with uncertainty—domains where purely symbolic engines face fundamental limitations?
    \item \textbf{RQ3:} How well does \method balance domain-specific performance enhancement with the maintenance of general capabilities?
\end{itemize}

We compared \method against several baselines: the unmodified Llama3-8B-Instruct model; the Llama3-8B-Instruct model fine-tuned using ground-truth reasoning trajectories (to ensure fair comparison, even though \method generates its own trajectories); and CARING \citep{yang2023neuro}, a method leveraging the Prolog engine for reasoning, which we consider the performance upper-bound for tasks fully amenable to symbolic deduction. The \method framework was tested in four configurations: full implementation with model averaging and multiple trajectories (MA+multiple), model averaging with a single trajectory (MA+single), multiple trajectories without model averaging (multiple), and a single trajectory without model averaging (single).

\subsection{Implementation Details}

\subsubsection{Models and Compute}
All experiments (trajectory generation, fine-tuning, and evaluation) were conducted on two NVIDIA A800 80GB graphics processing units (GPUs). We experiment with multiple backbone LLMs, including Llama3-8B-Instruct (\url{https://huggingface.co/meta-llama/Meta-Llama-3-8B-Instruct}) and {Qwen3-8B (\url{https://huggingface.co/Qwen/Qwen3-8B}).} Unless otherwise stated, we use the same prompting and evaluation protocol across backbones to ensure comparability.

\subsubsection{Fine-Tuning and Decoding Settings}
We perform continual supervised fine-tuning from each backbone checkpoint using LLaMA-Factory (\url{https://github.com/hiyouga/LLaMA-Factory}). We use Adam with a learning rate of $1\times10^{-6}$ for $3$ epochs. {To make the single-trajectory and multi-trajectory settings comparable despite different data volumes, we match the total number of optimization steps across settings.}

For generation during trajectory construction and evaluation, we use nucleus sampling with temperature $0.6$ and top\_p $0.9$ unless otherwise specified. {Notably, Qwen3 supports an internal ``thinking'' (built-in CoT) mode by default, whereas Llama3 does not. To keep the reasoning setting consistent across backbones (i.e., rely only on the external CoT prompting when used), we disable Qwen3's internal thinking mode by setting \texttt{enable\_thinking=False}. For model averaging, we use $\alpha=0.5$ to balance the base and fine-tuned checkpoints.}

{
\subsubsection{{Baselines}}

We compare against three families of baselines to cover complementary ways of improving (or outsourcing) reasoning: 
(i) \emph{training-free} methods that change prompting/decoding at inference time, 
(ii) \emph{training-based} methods that improve a backbone via supervised fine-tuning on reasoning trajectories, and 
(iii) \emph{LLMs-as-Agents} methods that translate inputs into executable symbolic programs and rely on an external solver.
Unless otherwise stated, training-free baselines follow the same task-specific prompting setup as in Table~\ref{tab:main_tractable}; \textbf{CoT-decoding} is the exception, as it elicits CoT through decoding and thus uses a basic prompt without explicit CoT exemplars.
\begin{itemize}
    \item \textbf{Base model.} The unmodified backbone (e.g., Llama3-8B-Instruct, Qwen3-8B) evaluated under the corresponding task prompts.

    \item \textbf{Self-Consistency \cite{wang2023selfconsistency}}: A decoding strategy for CoT prompting that samples multiple diverse reasoning paths and selects the most consistent final answer by aggregating across them (typically via majority voting).

    \item \textbf{CoT-decoding \cite{wang2024chain}}: Instead of injecting CoT via prompting, this method modifies the decoding process to surface latent CoT-like continuations (e.g., by exploring top-$k$ alternatives), and uses an internal confidence signal to prefer CoT paths. 
    Importantly, it does \emph{not} use explicit CoT exemplars in the prompt, so its prompting format differs from other training-free baselines.

    \item \textbf{Ground-truth training}: Supervised fine-tuning on dataset-provided reasoning trajectories (when available), followed by evaluation on the same tasks. 
    For QASC, ground-truth reasoning paths are not provided, so this baseline is not applicable (N/A).

    \item \textbf{Self-Taught Reasoner (STaR) \cite{zelikman2022star}}: A self-training loop that bootstraps CoT rationales: the model generates rationales, correct rationales are retained (or regenerated conditioned on the gold answer), and the model is fine-tuned on the resulting set. 
    In our setting, we use it as an alternative way to produce training trajectories and then fine-tune and evaluate the backbone.

    \item \textbf{CoT knowledge distillation \cite{magister-etal-2023-teaching}}: A distillation-style baseline where a student model is fine-tuned on CoT outputs generated by a stronger teacher, aiming to transfer reasoning behaviors to smaller models.

    \item \textbf{CARING (LLMs as Agents) \cite{yang2023neuro}}: An agentic symbolic baseline that first translates each problem into Prolog and then executes it with a Prolog engine, yielding both answers and interpretable proof logs. 
    On Prolog-tractable tasks, CARING serves as an approximate upper bound of pure symbolic reasoning; we therefore show it in grey and do not rank it against non-agent baselines.
\end{itemize}

}

\subsection{Dataset Selection and Rationale}
\label{Datasets}
Our dataset selection strategy aimed to evaluate \method across a spectrum of reasoning tasks, distinguishing between those solvable by pure symbolic methods and those requiring more nuanced neural capabilities. We categorized datasets as follows:

\subsubsection{Prolog-Tractable Datasets for In-Domain and Close-Reasoning Transfer Evaluation}
These datasets involve tasks where information is explicit and reasoning is largely deterministic, making them suitable for Prolog-based solutions:

\begin{itemize}
    \item \textbf{Grade School Math 8K (GSM8K)} \citep{cobbe2021training}: Comprises 8,500 high-quality grade school math word problems. We randomly selected 2,000 entries to generate reasoning trajectories (capped at 10 per problem, yielding 20,000 training instances), using the remainder for evaluation.
    
    \item \textbf{ProofWriter} \citep{d_proofwriter}: A logical reasoning dataset with knowledge bases and questions. We selected the most challenging open-world assumption (OWA) subset (482 knowledge bases, 10,190 questions), sampling 100 knowledge bases for trajectory generation (capped at 5 per question) and using the rest for evaluation.
    
    \item \textbf{Proof and Ontology-Generated Question-Answering (PrOntoQA)} \citep{saparov2023language}: A synthetic question-answering (QA) dataset for logical reasoning. We used the "Fictional Characters" depth-5 subset (500 pairs), selecting 100 for trajectory generation (capped at 5 per question) and the rest for evaluation.
    
    \item \textbf{Question Answering via Sentence Composition (QASC)} \citep{khot2020qasc}: A question-answering dataset where context sentences provide explicit facts needed to answer multi-hop questions. This dataset is well-suited for Prolog-based reasoning as all necessary information is provided explicitly. We used the official validation split for evaluation.
\end{itemize}

\subsubsection{Prolog-Intractable Datasets for Demonstrating Enhanced Intrinsic LLM Reasoning}
These datasets require reasoning with implicit knowledge, common sense, or uncertainty, posing challenges for purely symbolic systems:

\begin{itemize}
    \item \textbf{CommonsenseQA} \citep{talmor2018commonsenseqa}: A multiple-choice QA dataset requiring common-sense knowledge not explicitly stated in the prompt. Answering correctly often depends on unstated background assumptions that would be impractical to fully encode in a Prolog knowledge base.
    
    \item \textbf{StrategyQA} \citep{geva2021did}: A QA dataset where questions require multi-step reasoning under uncertainty. While some facts might be explicit, deriving the strategy often involves implicit knowledge or weighing evidence probabilistically—tasks fundamentally incompatible with classical Prolog's deterministic search paradigm.
\end{itemize}

\subsubsection{General OOD Datasets for Generalization Assessment}
These datasets assess broader capabilities and the impact of catastrophic forgetting:

\begin{itemize}
    \item \textbf{Massive Multitask Language Understanding (MMLU)} \citep{d_mmlu}: Encompasses 57 subjects, evaluating zero-shot and few-shot performance to gauge knowledge acquired during pretraining. Used entirely for evaluating generalization.
    
    \item \textbf{Graduate-Level Google-Proof Question Answering (GPQA)} \citep{d_gpqa}: Contains 448 expert-level multiple-choice questions in science, technology, engineering, and mathematics (STEM) fields. Used entirely to assess generalization.
    
    \item \textbf{HumanEval} \citep{d_humaneval}: Comprises 164 programming problems to evaluate code synthesis. Used entirely to assess OOD performance.
    
    \item \textbf{Mathematics Aptitude Test of Heuristics (MATH)} \citep{d_math}: Includes 12,500 challenging competition-level mathematics problems. Used to assess OOD performance in a complex reasoning domain.
\end{itemize}

{
\subsection{Evaluation Protocol}
\label{sec:eval_p}

Most of our benchmarks produce free-form
natural language outputs, including math (GSM8K),
classification-style reasoning (ProofWriter, PrOntoQA), and
multiple-choice QA (QASC, CommonsenseQA, StrategyQA, MMLU, GPQA).
On these tasks, exact string or field matching under-counts
correct predictions whenever the model paraphrases the reference
or wraps the answer in additional explanation. We therefore score
accuracy with GPT-4o as an LLM-as-judge. The judge sees the
question, the reference answer, and the model's output, and
returns a binary judgement on semantic alignment with the
reference. The same protocol is used for the industrial datasets
in Section~\ref{sec:indus_res}. HumanEval (pass@1) follows its respective standard
protocol. We manually verified a random subset of GPT-4o
judgements and observed high agreement.
}

\subsection{Experimental Results and Analysis}

\subsubsection{Performance on Prolog-Tractable Reasoning Tasks (RQ1)}
The performance of \method and baselines on Prolog-tractable tasks is presented in Table \ref{tab:main_tractable}.
{
As a prerequisite for leveraging Prolog-verified trajectories, one may reasonably worry that the overall pipeline could be bottlenecked by the LLM’s ability to translate natural-language problems into valid Prolog programs: if the transformation frequently fails (e.g., syntactically invalid code, malformed predicates, or inconsistent fact extraction), the resulting trajectory generation would become inefficient and costly, and the framework might only be practical when starting from already very strong base models. We therefore first quantify the reliability of the translation-and-execution stage on tasks with explicit logical structures that are fully tractable by symbolic deduction.

As shown in Table~\ref{tab:main_tractable}, even with relatively small 8B backbones (e.g., Llama3-8B and Qwen3-8B), CARING attains near-perfect accuracy on the Prolog-tractable benchmarks (100\% on ProofWriter and PrOntoQA, 97.6\% on QASC, and 98.1\% on GSM8K). On these tasks, Prolog is deterministic: once the LLM produces a correct, executable Prolog translation (facts/rules + query), the solver will return the right answer; when the translation is wrong or malformed, it will not. In this sense, CARING accuracy is a direct proxy for the end-to-end validity of the LLM-to-Prolog transformation, suggesting that invalid or unusable translations are rare (at most a few percent in the worst case). This also makes CARING an effective upper bound within the Prolog paradigm.

Across model families, \method consistently improves over the corresponding base model on Prolog-tractable benchmarks. As shown in Table~\ref{tab:main_tractable}, the gains hold not only for Llama3-8B-Instruct but also for Qwen3-8B: for instance, under the \texttt{multiple} configuration, \method reaches 87.81/98.19/100.00/96.70 on GSM8K/ProofWriter/PrOntoQA/QASC with Llama3-8B, and 95.54/99.75/100.00/98.30 with Qwen3-8B. Moreover, the improvements are not limited to a single training recipe---both model-averaged variants (MA+\texttt{single}/MA+\texttt{multiple}) and non-averaged variants (\texttt{single}/\texttt{multiple}) yield consistent gains, with \texttt{multiple} typically performing best. This pattern supports the intuition that exposing the model to a richer set of Prolog-verified trajectories strengthens domain-specific reasoning, while model averaging provides a stable option when one wants to trade a small amount of peak performance for better retention and robustness.

Notably, \method remains competitive against strong alternatives that also leverage multiple reasoning paths (e.g., Self-Consistency and other trajectory-based training baselines), suggesting that the advantage comes from the quality and verifiability of the generated trajectories rather than simply increasing the number of sampled rationales.

Notably, the reasoning trajectories generated by \method are logical and rigorously verified by Prolog, avoiding potential misinformation and maintaining consistent reasoning patterns. This provides an advantage over fine-tuning on potentially unverified ground-truth trajectories, as evidenced by \method (single) sometimes outperforming Llama3-ground-truth models. 

Figure \ref{fig:three_figures} illustrates that training with multiple reasoning trajectories per instruction yields substantial gains over a single trajectory, likely due to better modeling of multi-modal distributions inherent in LLM reasoning, thus enhancing robustness.

\begin{figure*}[!ht]
\centering
\subfloat[GSM8K]{\includegraphics[width=0.32\textwidth]{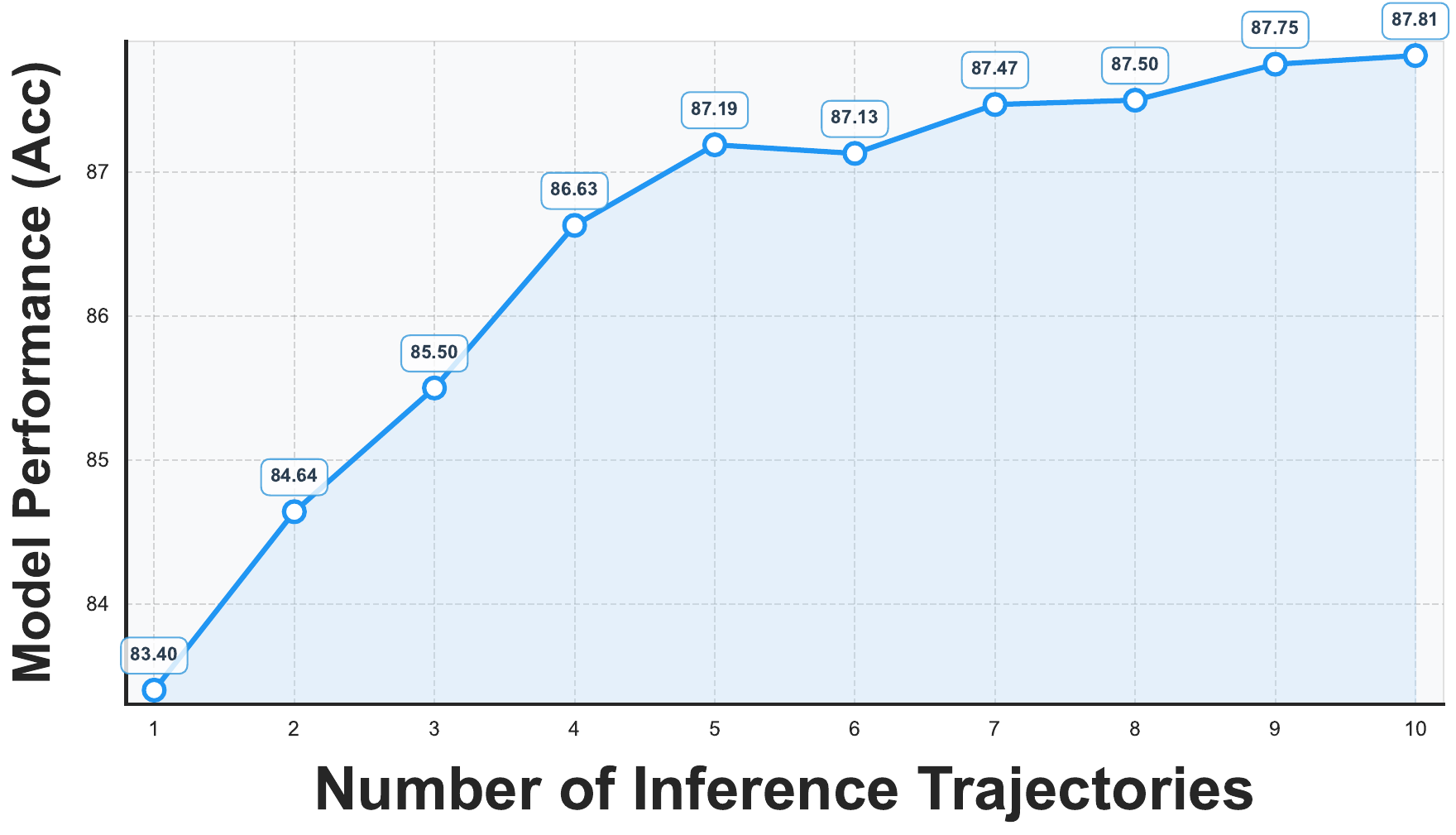}
}
\hfill
\subfloat[ProofWriter]{\includegraphics[width=0.32\textwidth]{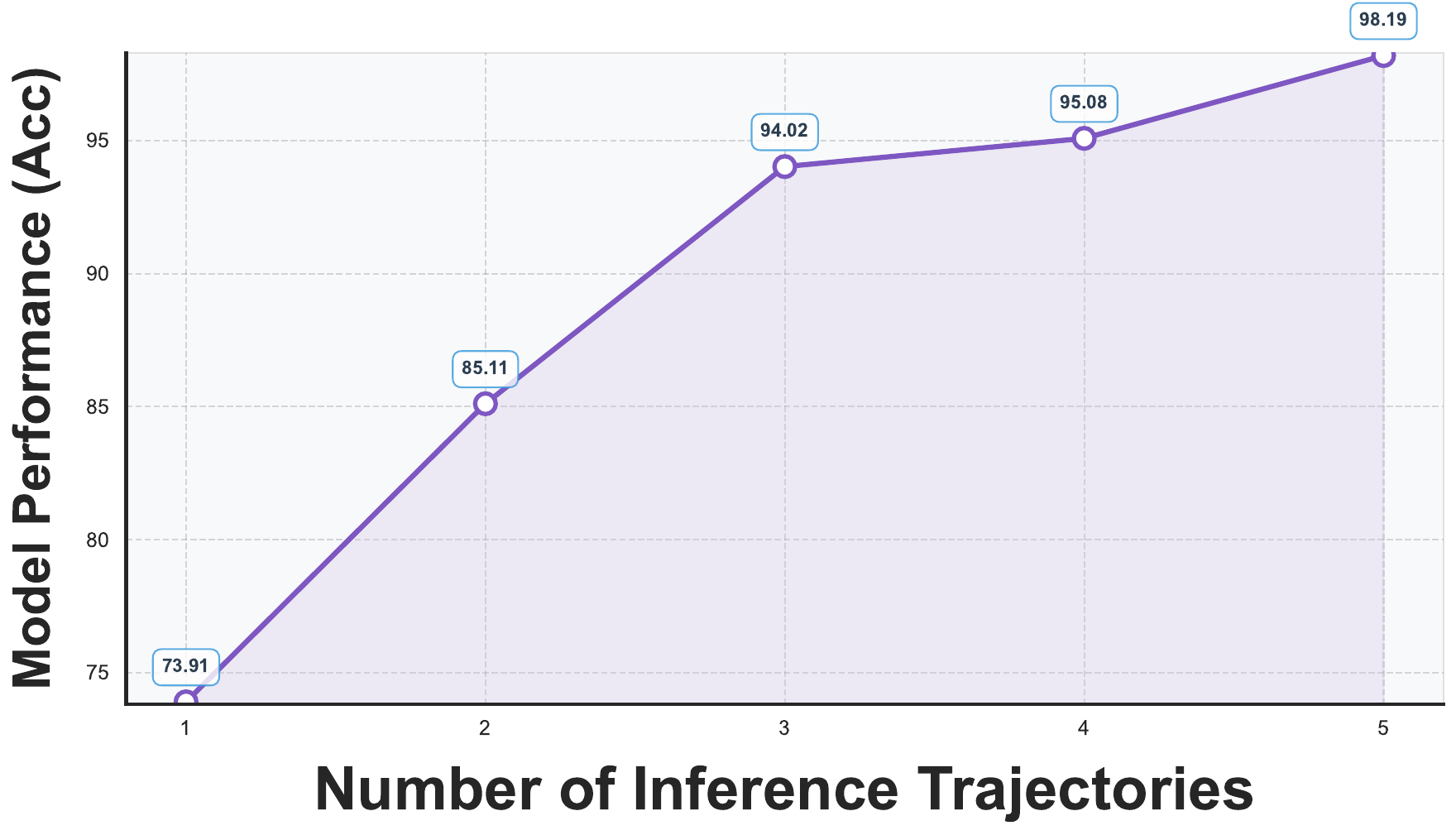}
}
\hfill
\subfloat[PrOntoQA]{\includegraphics[width=0.32\textwidth]{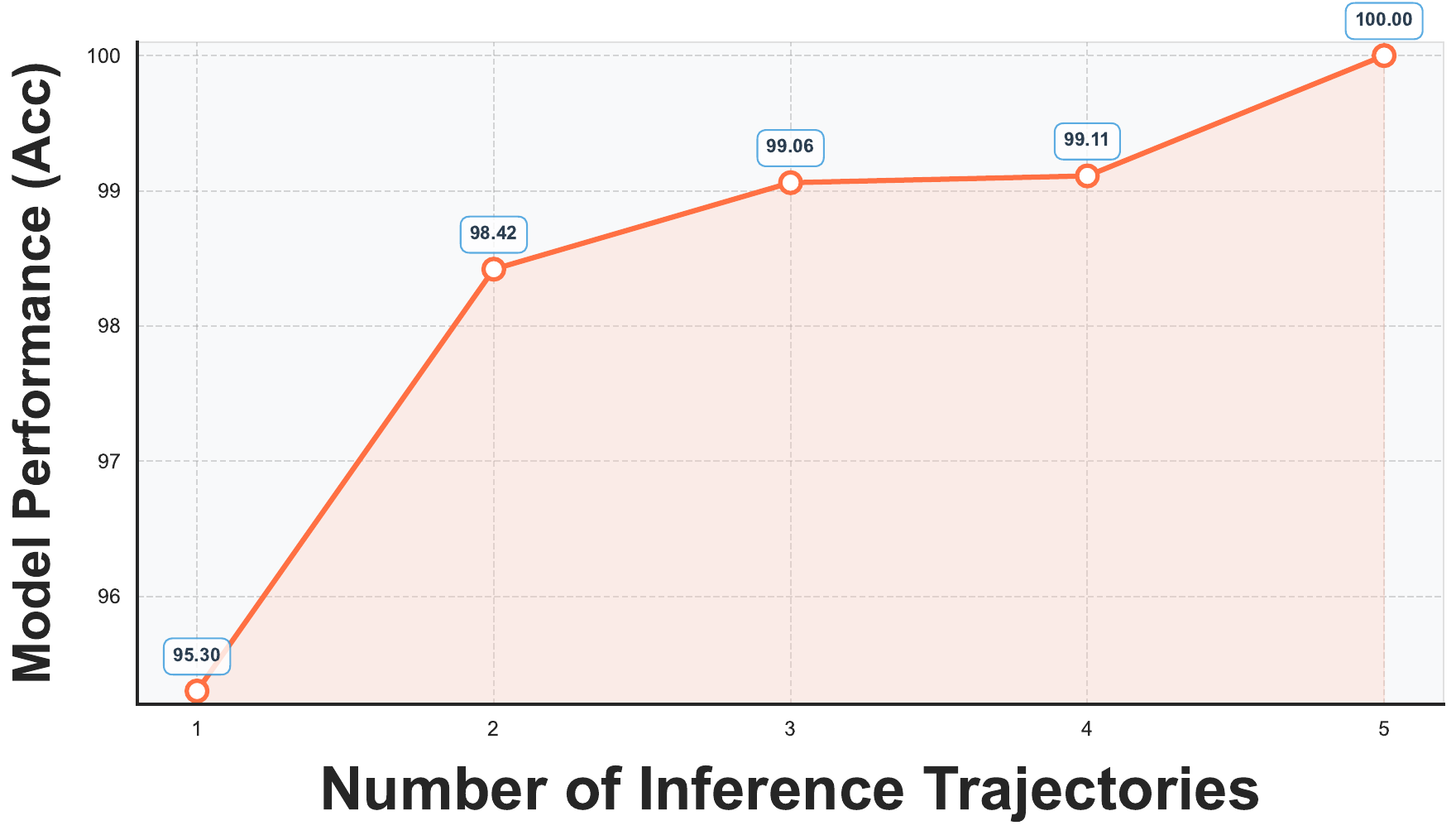}
}
\caption{Performance Evolution with Varying Numbers of Reasoning Trajectories: Showing enhanced logical reasoning capabilities and performance improvements through increased trajectory iterations.}
\label{fig:three_figures}
\end{figure*}

\subsubsection{Enhancing Intrinsic LLM Reasoning on Prolog-Intractable Tasks (RQ2)}
To demonstrate \method's unique value in improving the LLM's inherent reasoning capabilities, we evaluated it on tasks that are fundamentally challenging for purely symbolic systems like Prolog. These tasks require accessing implicit common-sense knowledge or reasoning under uncertainty, which Prolog's deterministic search over explicit knowledge bases cannot readily handle. Results are presented in Table \ref{tab:main_intractable}.

\begin{table}[t]
\centering
\caption{\textbf{Performance on Prolog-Intractable Reasoning Tasks.}}
\label{tab:main_intractable}
\setlength{\tabcolsep}{5pt}
\renewcommand{\arraystretch}{1.12}
\small
\begin{adjustbox}{width=0.95\columnwidth}
\begin{tabular}{lcc}
\toprule
\multirow{2}{*}{\textbf{Method}} & \multicolumn{2}{c}{\textbf{Acc $\uparrow$}} \\
\cmidrule(lr){2-3}
& \textbf{CommonsenseQA} & \textbf{StrategyQA} \\
\midrule
\rowcolor{gray!15}
\textbf{Prompt} & 0-shot & 0-shot \\
\midrule
Base model (Llama3-8B-Instruct) & 74.30 & 64.90 \\
\rowcolor{blue!10}
TLP (multiple) & \textbf{85.90} & \textbf{69.90} \\
CARING & \textcolor[HTML]{9B9B9B}{N/A} & \textcolor[HTML]{9B9B9B}{N/A} \\
\bottomrule
\end{tabular}
\end{adjustbox}

\vspace{1.2mm}
\begin{minipage}{0.98\linewidth}
\footnotesize
{\textbf{Notes.} `Acc' denotes accuracy. Best results are \textbf{bolded} and second-best are \underline{underlined}. CARING results are shown in grey and not ranked, as these tasks require implicit knowledge integration or probabilistic reasoning beyond the classical Prolog paradigm.}
\end{minipage}
\end{table}

Table \ref{tab:main_intractable} highlights a critical contribution of \method. For CommonsenseQA and StrategyQA, CARING is marked ``N/A" because these tasks represent theoretical barriers for classical Prolog, not implementation limitations. CommonsenseQA requires vast amounts of unstated background knowledge that would be impractical to fully encode in a Prolog knowledge base. Similarly, StrategyQA often involves weighing plausible alternatives or reasoning probabilistically—tasks fundamentally incompatible with Prolog's deterministic search paradigm.

\method significantly improves the Llama3-8B-Instruct model's performance on these tasks, with increases from 74.3\% to 85.9\% on CommonsenseQA and from 64.9\% to 69.9\% on StrategyQA using the ``multiple" configuration. This demonstrates that \method is not merely providing the LLM with an external symbolic tool; it is genuinely enhancing the model's intrinsic ability to perform complex reasoning by teaching it to internalize structured thought processes.

This capability to improve reasoning on tasks where symbolic engines cannot assist underscores the unique value of our framework compared to approaches that simply augment LLMs with external tools. By teaching the model itself to reason better through exposure to verified logical trajectories, \method enables knowledge transfer to contexts requiring implicit understanding or probabilistic reasoning.

\subsubsection{Impact of Model Averaging on Generalization and Catastrophic Forgetting (RQ3)}
Domain-specific training often leads to catastrophic forgetting of general capabilities. We employed model averaging techniques to mitigate this effect while preserving gains in specialized reasoning. Table \ref{tab:modelavg_ood} presents performance on general OOD tasks.

\begin{table*}[t]
\caption{\textbf{Impact of Model Averaging on General OOD Tasks.}}
\label{tab:modelavg_ood}
\vspace{2mm}
\centering
\setlength{\tabcolsep}{5.2pt}
\renewcommand{\arraystretch}{1.15}
\begin{adjustbox}{max width=\textwidth}
\begin{tabular}{llccccc}
\toprule
\multirow{2}{*}{\textbf{Method}} & \multirow{2}{*}{\textbf{Config.}} 
& \multicolumn{5}{c}{\textbf{Acc $\uparrow$}} \\
\cmidrule(lr){3-7}
& & \textbf{MMLU} & \textbf{GPQA} & \textbf{HumanEval} & \textbf{MATH} & \textbf{Avg.} \\
\midrule
\rowcolor{gray!15}
\textbf{Prompt} & -- & 5-shot & 0-shot & 0-shot & 4-shot, CoT & -- \\
\midrule
Base model & Llama3-8B-Instruct 
& \textbf{68.40} & \textbf{34.20} & \textbf{62.20} & 30.00 & \textbf{48.70} \\
\midrule
\rowcolor{gray!30} \multicolumn{7}{l}{\textbf{TLP variants (ours)}} \\
\rowcolor{blue!7}
TLP & MA+single   
& \underline{68.20} & \textbf{34.20} & \underline{62.10} & 29.30 & \underline{48.45} \\
\rowcolor{blue!7}
TLP & MA+multiple 
& 67.90 & \underline{33.80} & 61.90 & \textbf{30.20} & \underline{48.45} \\
\rowcolor{blue!7}
TLP & single      
& 65.40 & 32.80 & 60.20 & 29.20 & 46.90 \\
\rowcolor{blue!7}
TLP & multiple    
& 64.30 & 31.90 & 60.00 & \underline{30.10} & 46.58 \\
\bottomrule
\end{tabular}
\end{adjustbox}

\vspace{1.2mm}
\begin{minipage}{0.98\textwidth}
\footnotesize
{\textbf{Notes.} `Acc' denotes accuracy. Best results are \textbf{bolded} and second-best are \underline{underlined} (ties are \textbf{bolded}). `Avg.' is the mean over the four tasks. Model averaging (MA) largely preserves performance on general OOD tasks while still supporting the gains reported on specialized reasoning benchmarks elsewhere in the paper.}
\end{minipage}
\vspace{-2mm}
\end{table*}

As expected, the baseline Llama3-8B-Instruct model performs best on most general OOD tasks. However, model averaging within \method (MA+single, MA+multiple) effectively balances specialization and generalization. Compared to domain-specific training without averaging (single, multiple), the MA configurations show significantly better OOD performance with only a slight trade-off in specialized reasoning task performance (as seen in Table \ref{tab:main_tractable}). 

For example, MA+multiple achieved 67.9\% on MMLU and 33.8\% on GPQA, very close to baseline performance, while still enabling large improvements on reasoning tasks. This indicates that model averaging is a viable strategy to enhance robustness and generality. 

Intriguingly, performance on the MATH dataset remains relatively stable or even slightly improves (MA+multiple: 30.20\% vs. baseline: 30.00\%). This suggests that the arithmetic and logical reasoning skills cultivated by \method on structured tasks like GSM8K, ProofWriter, and PrOntoQA can generalize to complex MATH problems and resist catastrophic forgetting, likely because the reasoning patterns share structural similarities.
}

\subsection{Industrial Implementation of \method}
\label{sec:indus_res}

\begin{table}[t]
\centering
\caption{\textbf{Evaluation on Industrial Datasets.}}
\label{tab:industry}
\setlength{\tabcolsep}{6pt}
\renewcommand{\arraystretch}{1.12}
\small
\begin{adjustbox}{max width=\columnwidth}
\begin{tabular}{lcc}
\toprule
\multirow{2}{*}{\textbf{Method}} & \multicolumn{2}{c}{\textbf{Acc $\uparrow$}} \\
\cmidrule(lr){2-3}
& \textbf{Dataset 1} & \textbf{Dataset 2} \\
\midrule
\rowcolor{gray!15}
\textbf{Prompt} & \multicolumn{2}{c}{2-shot, CoT} \\
\midrule
Base (Llama3-8B-Instruct) & 76.90 & 52.80 \\
\rowcolor{blue!10}
TLP (multiple) & \textbf{93.50} & \textbf{94.70} \\
\bottomrule
\end{tabular}
\end{adjustbox}

\vspace{1.2mm}
\begin{minipage}{0.98\linewidth}
\footnotesize
{\textbf{Notes.} `Acc' denotes accuracy. Best results are \textbf{bolded} and second-best are \underline{underlined}.}
\end{minipage}
\end{table}

{

To evaluate the effectiveness of our proposed \method in
real-world production environments, we conducted experiments on
two financial reasoning scenarios using offline evaluation sets
curated from our production traffic. The first scenario focused
on a financial logic reasoning task, where the LLMs were required
to compute target metrics based on a provided context. This task
assessed the models' capabilities in computational reasoning and
mathematical operations to handle complex financial workflows
such as facility-fee accrual, covenant checks, and cash-flow
adjustments. Dataset~1 consists of $5{,}478$ training
cases and $1{,}000$ held-out test cases. The second scenario, a
security and compliance task, evaluated the models' ability to
detect security and compliance issues within the financial
domain, thereby measuring text-based logical reasoning
performance. Representative case types include sanctions
screening, transaction structuring detection, enhanced due
diligence triggers, and know-your-customer (KYC) refresh conditions. Dataset~2 consists of $6{,}107$ training cases and
$1{,}000$ held-out test cases. Both datasets contain a mix of
multiple-choice items and computational problems.

}

{
Both datasets are constructed from real, in-production cases at
our service rather than synthesized or web-scraped data, which we
view as essential for assessing the method under realistic
distribution shift, label noise, and case complexity. Personally
identifiable information was removed before any model
interaction, and all numerical figures shown in the synthetic
examples (Fig.~\ref{fig:industry_ex_logic},
Fig.~\ref{fig:industry_ex_compliance}) are fabricated. Each case
was annotated by in-house domain experts (financial analysts for
Dataset~1 and compliance officers for Dataset~2), with double
annotation and reconciliation on disagreement to control label
quality. The training and test splits do not overlap in customer
identifiers or case timestamps, so test cases were not seen
during trajectory generation or fine-tuning. We score accuracy on
the test sets using the GPT-4o LLM-as-judge protocol described in
Section~\ref{sec:eval_p}, which is appropriate for the
structured free-form outputs produced on these tasks, such as
target-metric dictionaries, action labels, and alert lists.}

\begin{figure}[t]
\vspace{-1mm}
\centering
\begin{exbox}{{Example A: Financial Logic Reasoning (computational)}}
\footnotesize
\textbf{Schema.} Facility terms + time-segmented balances + covenant definition.\\
\textbf{Context (abridged).}
Limit=10{,}000{,}000; APR=(4.20\%+2.10\%); undrawn fee=0.25\% (365-day).
Balances: 6.8M for days 1--19; 6.3M for days 20--30. letter of credit (LC)=1.2M; Lease=1.9M; earnings before interest, taxes, depreciation, and amortization (EBITDA)=3.1M; covenant max=3.0.\\
\textbf{Query.} Compute interest, undrawn fee, and covenant breach.\\
\textbf{Expected output.}
\begin{lstlisting}
interest=34268, undrawn_fee=696,
debt_ratio=3.03 -> breach=true
\end{lstlisting}
\textbf{Prolog sketch.}
\begin{lstlisting}
interest(P,Rate,Days,I):- I is P*Rate*Days/365.
debt_ratio(EndP,LC,Lease,EBITDA,R):- R is (EndP+LC+Lease)/EBITDA.
breach(R):- R>3.0.
\end{lstlisting}
\end{exbox}
\vspace{-2mm}
\caption{\textbf{Representative synthetic example} for the financial logic reasoning scenario. Values and entities are synthetic and do not correspond to any proprietary data.}
\label{fig:industry_ex_logic}
\vspace{-2mm}
\end{figure}

We benchmarked \method against a baseline model used in our service, Llama3-8B-Instruct, which was enhanced with two CoT task-specific example prompts. To optimize the reasoning capabilities of the LLM, we employed the ``Multiple'' setting with the number of reasoning trajectories set to 5. As shown in Table \ref{tab:industry}, \method significantly improved performance by enabling LLMs to utilize Prolog-generated, fully verified reasoning trajectories. Notably, the results indicate that financial text-based logical reasoning tasks demonstrated greater performance improvements with \method compared to mathematical reasoning tasks.

\begin{figure}[t]
\vspace{-1mm}
\centering
\begin{exbox}{{Example B: Security \& Compliance Reasoning (rule-based)}}
\footnotesize
\textbf{Schema.} Customer profile + transaction sequence + policy rules (time windows, thresholds, lists).\\
\textbf{Context (abridged).}
Risk=High; KYC last refreshed=2022-10-01.
Rules: (i) beneficiary in \{J-S\}$\Rightarrow$ \textsc{Block};
(ii) \textsc{Structuring}: in 24h, $\ge3$ transfers $<10k$ with sum$\ge20k$ $\Rightarrow$ \textsc{Review};
(iii) High-risk cross-border $\ge8k$ $\Rightarrow$ \textsc{Enhanced Due Diligence (EDD) Review};
(iv) KYC age$>24$ months $\Rightarrow$ \textsc{KYC Refresh}.\\
Tx: T1 9500 J-B; T2 9800 J-B; T3 9900 J-B (same day); T4 8200 J-C; T5 2000 J-S.\\
\textbf{Expected output.}
\begin{lstlisting}
blocked=[T5], alerts=[SANCTIONS_MATCH,
STRUCTURING, EDD, KYC_REFRESH],
case_action=BLOCK
\end{lstlisting}
\textbf{Prolog sketch.}
\begin{lstlisting}
block(T):- tx(T,_,_,J), sanctioned(J).
structuring:- count_small_24h(N), sum_small_24h(S), N>=3, S>=20000.
edd:- risk(high), exists_xborder_ge(8000).
kyc_refresh:- kyc_age_months(M), M>24.
\end{lstlisting}
\end{exbox}
\vspace{-2mm}
\caption{\textbf{Representative synthetic example} for the security \& compliance scenario. Values and entities are synthetic and do not correspond to any proprietary data.}
\label{fig:industry_ex_compliance}
\vspace{-2mm}
\end{figure}

From an industry perspective, \method not only enhances LLM performance but also provides a robust solution for real-world applications in the financial reasoning scenario. Validation with actual business data confirms the approach’s reliability and effectiveness in solving complex financial problems. This establishes a practical and scalable technological pathway for the intelligent transformation of financial institutions, offering actionable insights and a competitive edge in financial reasoning and compliance tasks.

\subsection{Discussion of Experimental Findings}
Our experimental results collectively demonstrate the multifaceted benefits of the \method framework:

First, on tasks amenable to symbolic deduction (GSM8K, ProofWriter, PrOntoQA, QASC), \method enables the LLM to learn from Prolog-verified reasoning trajectories, achieving performance levels that significantly surpass baseline LLMs and approach the theoretical limits set by pure symbolic engines like CARING. The use of multiple, validated trajectories proves particularly effective in enhancing this learning process.

Second, and critically, \method extends the LLM's capabilities beyond the confines of purely symbolic reasoning. On tasks requiring implicit common-sense knowledge (CommonsenseQA) or reasoning under uncertainty (StrategyQA)—where symbolic systems like CARING are fundamentally inapplicable—\method substantially improves the LLM's performance. This indicates a genuine enhancement of the model's intrinsic reasoning faculties, teaching it how to reason more effectively rather than merely providing an external symbolic tool.

Finally, the incorporation of model averaging techniques within \method offers a practical solution to the challenge of catastrophic forgetting. It allows for a favorable balance between achieving high performance on specialized reasoning tasks and maintaining strong generalization capabilities across diverse out-of-distribution benchmarks. The observed resilience and even slight improvement on the MATH dataset further suggest that the reasoning skills imparted by \method possess meaningful transferability to complex, unseen reasoning problems.

In summary, \method presents a robust framework for improving LLM reasoning by synergizing the precision of symbolic logic with the flexible, knowledge-rich nature of neural models, addressing both explicit deductive tasks and more nuanced reasoning challenges that require implicit knowledge or probabilistic thinking.

\section{Conclusion} 

In this paper, we introduce \method, a framework designed to enhance the logical reasoning of LLMs by imitating Prolog-based strictly logical reasoning trajectories in a self-driven manner. The design is \textbf{simple, straightforward, and effective}, ensuring ease of use in industrial applications. Our work is not aimed at \textbf{securing state-of-the-art (SOTA) results} on existing task-specific evaluation datasets.  Our focus lies in \textbf{improving the internal reasoning capability of LLMs.} Specifically, by correcting and learning on multiple reasoning trajectories, we can enhance the model's reasoning capacity for specific tasks. Our initial findings show improved task performance on both specialized reasoning and general benchmarks. We also evaluate \method in real-world financial reasoning scenarios which demonstrate superior performance compared with ICL baseline methods.  This foundational framework will be further developed and extensively evaluated across various tasks to refine its methodology in future works.

{
\section{Limitations}
\label{sec:limitations}

The main limitation of this work concerns what we mean by improved
reasoning. Our evidence is behavioral. Higher accuracy on reasoning
benchmarks after training on Prolog-verified trajectories is consistent
with stronger reasoning competence, but it is also consistent with the
model having learned a cleaner output format rather than a deeper change
in how it arrives at its answers. The same ambiguity has been noted for
chain-of-thought methods more broadly, where the generated rationale
need not faithfully reflect the underlying computation
\cite{turpin2023language, nye2022show}. We therefore use ``intrinsic
reasoning'' in this paper to mean that the model reaches correct
answers without calling an external solver at inference time, rather
than as a claim about any deeper mechanism inside the model. Separating
style acquisition from a genuine change in reasoning behavior is a
question we cannot fully settle here, and we treat it as open.

A related limitation lies in the scope of the supervision signal. Our
training trajectories come, by construction, from problems whose answers
can be verified by Prolog, which biases the training distribution
toward tasks with explicit logical or arithmetic structure. The transfer
we observe to CommonsenseQA and StrategyQA is encouraging but empirical.
We do not have a theoretical account of when reasoning learned from
verifiable problems should generalize to tasks that rely on implicit
knowledge or probabilistic judgement, and we cannot rule out weaker
transfer on benchmarks farther from the training distribution we used.
}

\bibliographystyle{IEEEtranN}
\bibliography{TCDS_Arxiv_References}

@inproceedings{7780381,
  title = {Neural Module Networks},
  author = {Jacob Andreas and Marcus Rohrbach and Trevor Darrell and Dan Klein},
  year = {2016},
  booktitle = {Proc. of CVPR},
  pages = {39--48}
}

@article{achiam2023gpt,
  title = {{GPT-4} technical report},
  author = {Achiam, Josh and others},
  year = {2023},
  journal = {arXiv preprint arXiv:2303.08774}
}

@inproceedings{besta2024graph,
  title = {Graph of Thoughts: Solving Elaborate Problems with Large Language Models},
  author = {Maciej Besta and others},
  year = {2024},
  booktitle = {Proc. of AAAI},
  pages = {17682--17690}
}

@inproceedings{bowen1979prolog,
  title = {Prolog},
  author = {Bowen, Kenneth A},
  year = {1979},
  booktitle = {Proceedings of the 1979 annual conference},
  pages = {14--23}
}

@article{bubeck2023sparks,
  title = {Sparks of artificial general intelligence: Early experiments with {GPT-4}},
  author = {Bubeck, S{\'e}bastien and others},
  year = {2023},
  journal = {arXiv preprint arXiv:2303.12712}
}

@article{casper2023open,
  title = {Open problems and fundamental limitations of reinforcement learning from human feedback},
  author = {Casper, Stephen and others},
  year = {2023},
  journal = {Transactions on Machine Learning Research}
}

@article{chen2023challenges,
  title = {Challenges and Contributing Factors in the Utilization of Large Language Models ({LLMs})},
  author = {Chen, Xiaoliang and others},
  year = {2023},
  journal = {arXiv preprint arXiv:2310.13343}
}

@article{cobbe2021training,
  title = {Training verifiers to solve math word problems},
  author = {Cobbe, Karl and others},
  year = {2021},
  journal = {arXiv preprint arXiv:2110.14168}
}

@inproceedings{creswell2023selectioninference,
  title = {Selection-Inference: Exploiting Large Language Models for Interpretable Logical Reasoning},
  author = {Antonia Creswell and Murray Shanahan and Irina Higgins},
  year = {2023},
  booktitle = {Proc. of ICLR}
}

@inproceedings{d_gpqa,
  title = {{GPQA}: A graduate-level google-proof q\&a benchmark},
  author = {Rein, David and others},
  year = {2024},
  booktitle = {Proc. of COLM}
}

@article{d_humaneval,
  title = {Evaluating large language models trained on code},
  author = {Chen, Mark and others},
  year = {2021},
  journal = {arXiv preprint arXiv:2107.03374}
}

@inproceedings{d_math,
  title = {Measuring mathematical problem solving with the {MATH} dataset},
  author = {Hendrycks, Dan and others},
  year = {2021},
  booktitle = {Proc. of NeurIPS Datasets and Benchmarks}
}

@inproceedings{d_mmlu,
  title = {Measuring Massive Multitask Language Understanding},
  author = {Dan Hendrycks and others},
  year = {2021},
  booktitle = {Proc. of ICLR}
}

@inproceedings{d_proofwriter,
  title = {{P}roof{W}riter: Generating Implications, Proofs, and Abductive Statements over Natural Language},
  author = {Tafjord, Oyvind  and Dalvi, Bhavana  and Clark, Peter},
  year = {2021},
  booktitle = {Proc. of ACL Findings},
  pages = {3621--3634}
}

@misc{deng2024promotingequalitylargelanguage,
  title = {Promoting Equality in Large Language Models: Identifying and Mitigating the Implicit Bias based on Bayesian Theory},
  author = {Yongxin Deng and others},
  year = {2024},
  journal = {arXiv preprint arXiv:2408.10608}
}

@inproceedings{dong2022survey,
  title = {A survey on in-context learning},
  author = {Dong, Qingxiu and others},
  year = {2024},
  booktitle = {Proc. of EMNLP},
  pages = {1107--1128}
}

@inproceedings{dong2023abilities,
  title = {How abilities in large language models are affected by supervised fine-tuning data composition},
  author = {Dong, Guanting and others},
  year = {2024},
  booktitle = {Proc. of ACL},
  pages = {177--198}
}

@article{drori2022neural,
  title = {A neural network solves, explains, and generates university math problems by program synthesis and few-shot learning at human level},
  author = {Drori, Iddo and others},
  year = {2022},
  journal = {Proceedings of the National Academy of Sciences},
  pages = {e2123433119}
}

@article{dubey2024llama,
  title = {The {Llama 3} herd of models},
  author = {Dubey, Abhimanyu and others},
  year = {2024},
  journal = {arXiv preprint arXiv:2407.21783}
}

@book{fagin2004reasoning,
  title = {Reasoning about knowledge},
  author = {Fagin, Ronald and Halpern, Joseph Y and Moses, Yoram and Vardi, Moshe},
  year = {2004},
  publisher = {MIT press}
}

@inproceedings{fu2022complexity,
  title = {Complexity-Based Prompting for Multi-step Reasoning},
  author = {Yao Fu and Hao Peng and Ashish Sabharwal and Peter Clark and Tushar Khot},
  year = {2023},
  booktitle = {Proc. of ICLR}
}

@article{geva2021did,
  title = {Did Aristotle Use a Laptop? A Question Answering Benchmark with Implicit Reasoning Strategies},
  author = {Geva, Mor  and Khashabi, Daniel  and Segal, Elad  and Khot, Tushar  and Roth, Dan  and Berant, Jonathan},
  year = {2021},
  journal = {TACL},
  pages = {346--361}
}

@inproceedings{gunel2020supervised,
  title = {Supervised Contrastive Learning for Pre-trained Language Model Fine-tuning},
  author = {Beliz Gunel and Jingfei Du and Alexis Conneau and Veselin Stoyanov},
  year = {2021},
  booktitle = {Proc. of ICLR}
}

@inproceedings{gupta2019neural,
  title = {Neural Module Networks for Reasoning over Text},
  author = {Nitish Gupta and Kevin Lin and Dan Roth and Sameer Singh and Matt Gardner},
  year = {2020},
  booktitle = {Proc. of ICLR}
}

@inproceedings{hudson2019learning,
  title = {Learning by Abstraction: The Neural State Machine},
  author = {Drew A. Hudson and Christopher D. Manning},
  year = {2019},
  booktitle = {Proc. of NeurIPS},
  pages = {5901--5914}
}

@article{hussein2017imitation,
  title = {Imitation learning: A survey of learning methods},
  author = {Hussein, Ahmed and Gaber, Mohamed Medhat and Elyan, Eyad and Jayne, Chrisina},
  year = {2017},
  journal = {ACM Computing Surveys},
  pages = {21:1--21:35},
  volume = {50},
  number = {2}
}

@inproceedings{kazemi-etal-2023-lambada,
  title = {{LAMBADA}: Backward Chaining for Automated Reasoning in Natural Language},
  author = {Kazemi, Mehran  and Kim, Najoung  and Bhatia, Deepti  and Xu, Xin  and Ramachandran, Deepak},
  year = {2023},
  booktitle = {Proc. of ACL},
  pages = {6547--6568}
}

@inproceedings{khot2020qasc,
  title = {{QASC:} {A} Dataset for Question Answering via Sentence Composition},
  author = {Tushar Khot and Peter Clark and Michal Guerquin and Peter Jansen and Ashish Sabharwal},
  year = {2020},
  booktitle = {Proc. of AAAI},
  pages = {8082--8090}
}

@inproceedings{kojima2022large,
  title = {Large Language Models are Zero-Shot Reasoners},
  author = {Takeshi Kojima and Shixiang Shane Gu and Machel Reid and Yutaka Matsuo and Yusuke Iwasawa},
  year = {2022},
  booktitle = {Proc. of NeurIPS},
  pages = {22199--22213}
}

@inproceedings{lee2022factuality,
  title = {Factuality Enhanced Language Models for Open-Ended Text Generation},
  author = {Nayeon Lee and others},
  year = {2022},
  booktitle = {Proc. of NeurIPS}
}

@inproceedings{lester2021power,
  title = {The Power of Scale for Parameter-Efficient Prompt Tuning},
  author = {Lester, Brian  and Al-Rfou, Rami  and Constant, Noah},
  year = {2021},
  booktitle = {Proc. of EMNLP},
  pages = {3045--3059}
}

@inproceedings{li2021prefix,
  title = {Prefix-Tuning: Optimizing Continuous Prompts for Generation},
  author = {Li, Xiang Lisa  and Liang, Percy},
  year = {2021},
  booktitle = {Proc. of ACL},
  pages = {4582--4597}
}

@article{lin2023speciality,
  title = {Speciality vs generality: An empirical study on catastrophic forgetting in fine-tuning foundation models},
  author = {Lin, Yong and others},
  year = {2023},
  journal = {arXiv preprint arXiv:2309.06256}
}

@article{long2023large,
  title = {Large language model guided tree-of-thought},
  author = {Long, Jieyi},
  year = {2023},
  journal = {arXiv preprint arXiv:2305.08291}
}

@inproceedings{lyu2023faithful,
  title = {Faithful Chain-of-Thought Reasoning},
  author = {Lyu, Qing and others},
  year = {2023},
  booktitle = {Proc. of AACL},
  pages = {305--329}
}

@inproceedings{magister-etal-2023-teaching,
  title = {Teaching Small Language Models to Reason},
  author = {Magister, Lucie Charlotte  and Mallinson, Jonathan  and Adamek, Jakub  and Malmi, Eric  and Severyn, Aliaksei},
  year = {2023},
  booktitle = {Proc. of ACL},
  pages = {1773--1781}
}

@inproceedings{neelakantan2017learning,
  title = {Learning a Natural Language Interface with Neural Programmer},
  author = {Arvind Neelakantan and Quoc V. Le and Mart{\'{\i}}n Abadi and Andrew McCallum and Dario Amodei},
  year = {2017},
  booktitle = {Proc. of ICLR}
}

@article{nye2022show,
  title = {Show Your Work: Scratchpads for Intermediate Computation with Language Models},
  author = {Maxwell Nye and others},
  year = {2021},
  journal = {arXiv preprint arXiv:2112.00114}
}

@article{osa2018algorithmic,
  title = {An algorithmic perspective on imitation learning},
  author = {Takayuki Osa and Joni Pajarinen and Gerhard Neumann and J. Andrew Bagnell and Pieter Abbeel and Jan Peters},
  year = {2018},
  journal = {Foundations and Trends in Robotics},
  pages = {1--179},
  volume = {7},
  number = {1--2}
}

@inproceedings{pan2023logic,
  title = {Logic-{LM}: Empowering Large Language Models with Symbolic Solvers for Faithful Logical Reasoning},
  author = {Pan, Liangming  and Albalak, Alon  and Wang, Xinyi  and Wang, William},
  year = {2023},
  booktitle = {Proc. of EMNLP Findings},
  pages = {3806--3824}
}

@inproceedings{ribeiro2023street,
  title = {{STREET:} {A} Multi-Task Structured Reasoning and Explanation Benchmark},
  author = {Danilo Neves Ribeiro and others},
  year = {2023},
  booktitle = {Proc. of ICLR}
}

@inproceedings{saparov2023language,
  title = {Language Models Are Greedy Reasoners: {A} Systematic Formal Analysis of Chain-of-Thought},
  author = {Abulhair Saparov and He He},
  year = {2023},
  booktitle = {Proc. of ICLR}
}

@article{shi2024continual,
  title = {Continual Learning of Large Language Models: A Comprehensive Survey},
  author = {Shi, Haizhou and others},
  year = {2026},
  journal = {ACM Computing Surveys},
  volume = {58},
  number = {5},
  pages = {120:1--120:42}
}

@inproceedings{shin2020autoprompt,
  title = {{A}uto{P}rompt: {E}liciting {K}nowledge from {L}anguage {M}odels with {A}utomatically {G}enerated {P}rompts},
  author = {Shin, Taylor  and Razeghi, Yasaman  and Logan IV, Robert L.  and Wallace, Eric  and Singh, Sameer},
  year = {2020},
  booktitle = {Proc. of EMNLP},
  pages = {4222--4235}
}

@inproceedings{tafjord2022entailer,
  title = {Entailer: Answering Questions with Faithful and Truthful Chains of Reasoning},
  author = {Tafjord, Oyvind  and Dalvi Mishra, Bhavana  and Clark, Peter},
  year = {2022},
  booktitle = {Proc. of EMNLP},
  pages = {2078--2093}
}

@inproceedings{talmor2018commonsenseqa,
  title = {{C}ommonsense{QA}: A Question Answering Challenge Targeting Commonsense Knowledge},
  author = {Talmor, Alon  and Herzig, Jonathan  and Lourie, Nicholas  and Berant, Jonathan},
  year = {2019},
  booktitle = {Proc. of NAACL},
  pages = {4149--4158}
}

@inproceedings{turpin2023language,
  title = {Language Models Don't Always Say What They Think: Unfaithful Explanations in Chain-of-Thought Prompting},
  author = {Miles Turpin and Julian Michael and Ethan Perez and Samuel R. Bowman},
  year = {2023},
  booktitle = {Proc. of NeurIPS},
  pages = {74952--74965}
}

@article{varshney2023stitch,
  title = {A stitch in time saves nine: Detecting and mitigating hallucinations of {LLMs} by validating low-confidence generation},
  author = {Varshney, Neeraj and Yao, Wenlin and Zhang, Hongming and Chen, Jianshu and Yu, Dong},
  year = {2023},
  journal = {arXiv preprint arXiv:2307.03987}
}

@inproceedings{wang2023selfconsistency,
  title = {Self-Consistency Improves Chain of Thought Reasoning in Language Models},
  author = {Xuezhi Wang and others},
  year = {2023},
  booktitle = {Proc. of ICLR}
}

@inproceedings{wang2024chain,
  title = {Chain-of-Thought Reasoning Without Prompting},
  author = {Xuezhi Wang and Denny Zhou},
  year = {2024},
  booktitle = {Proc. of NeurIPS},
  pages = {66383--66409}
}

@inproceedings{wei2022chain,
  title = {Chain-of-Thought Prompting Elicits Reasoning in Large Language Models},
  author = {Jason Wei and others},
  year = {2022},
  booktitle = {Proc. of NeurIPS}
}

@article{wielemaker2012swi,
  title = {{SWI-Prolog}},
  author = {Wielemaker, Jan and Schrijvers, Tom and Triska, Markus and Lager, Torbj{\"o}rn},
  year = {2012},
  journal = {Theory and Practice of Logic Programming},
  pages = {67--96},
  volume = {12},
  number = {1--2}
}

@inproceedings{workrethinking,
  title = {Rethinking the Role of Demonstrations: What Makes In-Context Learning Work?},
  author = {Min, Sewon and others},
  year = {2022},
  booktitle = {Proc. of EMNLP},
  pages = {11048--11064}
}

@book{wos1992automated,
  title = {Automated reasoning introduction and applications},
  author = {Wos, Larry and Overbeek, Ross and Lusk, Ewing and Boyle, Jim},
  year = {1992},
  publisher = {McGraw-Hill, Inc.}
}

@article{wu2023comparative,
  title = {A comparative study of open-source large language models, {GPT-4} and claude 2: Multiple-choice test taking in nephrology},
  author = {Wu, Sean and others},
  year = {2023},
  journal = {arXiv preprint arXiv:2308.04709}
}

@inproceedings{yang2023neuro,
  title = {Neuro-symbolic integration brings causal and reliable reasoning proofs},
  author = {Yang, Sen and Li, Xin and Cui, Leyang and Bing, Lidong and Lam, Wai},
  year = {2025},
  booktitle = {Proc. of NAACL Findings},
  pages = {5747--5759}
}

@inproceedings{yao2024tree,
  title = {Tree of Thoughts: Deliberate Problem Solving with Large Language Models},
  author = {Shunyu Yao and others},
  year = {2023},
  booktitle = {Proc. of NeurIPS},
  pages = {11809--11822}
}

@inproceedings{ye2022the,
  title = {The Unreliability of Explanations in Few-shot Prompting for Textual Reasoning},
  author = {Xi Ye and Greg Durrett},
  year = {2022},
  booktitle = {Proc. of NeurIPS}
}

@inproceedings{zelikman2022star,
  title = {{STaR}: Bootstrapping Reasoning With Reasoning},
  author = {Eric Zelikman and Yuhuai Wu and Jesse Mu and Noah D. Goodman},
  year = {2022},
  booktitle = {Proc. of NeurIPS},
  pages = {15476--15488}
}

@inproceedings{zhai2023investigating,
  title = {Investigating the Catastrophic Forgetting in Multimodal Large Language Model Fine-Tuning},
  author = {Zhai, Yuexiang and others},
  year = {2024},
  booktitle = {Proc. of CPAL},
  pages = {202--227}
}

@inproceedings{zhang2022automatic,
  title = {Automatic Chain of Thought Prompting in Large Language Models},
  author = {Zhuosheng Zhang and Aston Zhang and Mu Li and Alex Smola},
  year = {2023},
  booktitle = {Proc. of ICLR}
}

@article{zhang2023hallucination,
  title = {Siren's Song in the AI Ocean: A Survey on Hallucination in Large Language Models},
  author = {Zhang, Yue and others},
  year = {2023},
  journal = {arXiv preprint arXiv:2309.01219}
}

@article{zhang2024cumulative,
  title = {Cumulative reasoning with large language models},
  author = {Zhang, Yifan and Yang, Jingqin and Yuan, Yang and Yao, Andrew Chi-Chih},
  year = {2025},
  journal = {Transactions on Machine Learning Research}
}

@article{zhao2023survey,
  title = {A survey of large language models},
  author = {Zhao, Wayne Xin and others},
  year = {2023},
  journal = {arXiv preprint arXiv:2303.18223}
}

@article{zhao2024marco,
  title = {{Marco-o1}: Towards open reasoning models for open-ended solutions},
  author = {Zhao, Yu and others},
  year = {2024},
  journal = {arXiv preprint arXiv:2411.14405}
}

@inproceedings{zhou2022large,
  title = {Large Language Models are Human-Level Prompt Engineers},
  author = {Yongchao Zhou and others},
  year = {2023},
  booktitle = {Proc. of ICLR}
}

@article{zhou2024survey,
  title = {A survey on efficient inference for large language models},
  author = {Zhou, Zixuan and others},
  year = {2024},
  journal = {arXiv preprint arXiv:2404.14294}
}

\clearpage

\appendices

{
\section{Prolog Representation: Facts, Rules, and Targets}
\label{app:prolog_rep}

\subsection{Terminology and Mapping}
\label{app:prolog_terms}
To make the Prolog interface concrete, we clarify the roles of \emph{facts}, \emph{rules}, and \emph{targets} used throughout the paper.
Intuitively, facts encode instance-specific information from the input, rules encode reusable domain constraints, and targets specify what we want the solver to derive.

\begin{exbox}{{Concepts used in our Prolog interface (plain-language view)}}
\small
\centering
\begin{tabular}{@{}p{0.16\linewidth}p{0.78\linewidth}@{}}
\toprule
\textbf{Term} & \textbf{What it refers to in our framework} \\
\midrule
Facts & Instance-specific statements extracted from the input (entities, attributes, relations), represented as Prolog predicates. \\
Rules & Reusable logical constraints or domain rules (implications) that, together with facts, allow symbolic derivation. \\
Targets & The quantities/labels to be derived (e.g., the final answer, a classification label, or an intermediate decision), expressed as a Prolog query. \\
Query & The executable Prolog goal that asks for the target (e.g., \texttt{answer(X).} or \texttt{breach.}). \\
Proof / Trace & The Prolog derivation steps (search log) that justify why the query succeeds and how variables are instantiated. \\
\bottomrule
\end{tabular}
\end{exbox}

\subsection{Illustrative End-to-End Example}
\label{app:prolog_example}
We provide a minimal but representative example showing how a natural-language instance is translated into Prolog facts/rules and a target query, and how the solver output is mapped back to the final answer.

\begin{exbox}{{Example: rule-based decision with explicit targets}}
\footnotesize
\textbf{Natural-language instance.}
\emph{A company grants discounts under the following policy: Gold members get a 10\% discount. If a customer has spent at least \$1000 and is not flagged for fraud, they become Gold. Alice spent \$1200 and is not flagged. What discount does Alice get?}

\vspace{0.8mm}
\textbf{Facts (instance-specific).}
\begin{lstlisting}
spent(alice, 1200).
flagged(alice, no).
\end{lstlisting}

\textbf{Rules (domain constraints).}
\begin{lstlisting}
gold(U) :- spent(U, S), S >= 1000, flagged(U, no).
discount(U, 10) :- gold(U).
discount(U, 0) :- \+ gold(U).
\end{lstlisting}

\textbf{Target (query).} We ask the solver to derive the target discount:
\begin{lstlisting}
?- discount(alice, D).
\end{lstlisting}

\textbf{Solver output (answer instantiation).}
\begin{lstlisting}
D = 10.
\end{lstlisting}

\textbf{Mapping back.} The system converts \texttt{D=10} into the final answer: \emph{Alice gets a 10\% discount.}
\end{exbox}

\subsection{Validation and Filtering in Trajectory Generation}
\label{app:prolog_filter}
In our pipeline, a generated trajectory is retained only if the Prolog program is (i) syntactically valid and executable, and (ii) semantically consistent in the sense that the target query can be evaluated without missing predicates or ill-typed arguments. Cases that fail parsing/execution (e.g., malformed predicates, missing targets, or inconsistent fact extraction) are filtered out and not used for training. This makes the retained trajectories fully verifiable by construction.
}

\section{Prompt demo}
\label{Prompt demo}
In this section, we present the few-shot demonstration prompt utilized. It should be noted that due to space constraints, all prompts are displayed in their 1-shot condition, wherein content preceding the \textcolor{red}{red hash line} constitutes the input, and subsequent content represents the large language model's (LLM's) response.\par

\begin{mdframed}[
    linewidth=1pt,
    roundcorner=10pt,
    backgroundcolor=green!5,
    innerleftmargin=15pt,
    innerrightmargin=15pt,
    innertopmargin=15pt,
    innerbottommargin=15pt
]
The prompt demonstration of $\text{prompt}_{rft}$\\[0.5em]

\underline{System:} \\[0.5em]
\small
You are a helpful assistant who \textbf{produces Prolog code} to solve problems.\\[1em]

\underline{User:}\\[0.5em]
\small
Could you please help me write Prolog code to solve the following arithmetic reasoning problem? You should use consistent variable names for coreferent entities or attributes throughout the code. Start by coding the given context after the "/* Context */" comment. Then code the query that represents the question after the "/* Query */" comment.\\[0.5em]

Here is the problem:\\
sent-1: Tina makes \$18.00 an hour.\\
sent-2: If she works more than 8 hours per shift,\\
sent-3: she is eligible for overtime,\\
sent-4: which is paid by your hourly wage + 1/2 your hourly wage.\\
sent-5: If she works 10 hours every day for 5 days,\\
sent-6: how much money does she make?\\[1em]

\underline{Assistant:}\\[0.5em]
\small
I am happy to help you write Prolog code to solve this arithmetic reasoning problem. Here is the Prolog code:\\[0.5em]

\begin{lstlisting}[
    language=Prolog,
    basicstyle=\tiny\ttfamily,  
    breaklines=true,
    columns=flexible,
    frame=single,
    backgroundcolor=\color{gray!5},
    showstringspaces=false,
    numbers=none
]
/* Context */

wage(18.00).

overtime_wage(W) :- 
    wage(W1), 
    W is 1.5 * W1.

regular_earning_for_day(E) :- 
    wage(W),
    E is 8 * W.

overtime_hours(H) :-
    H is 10 - 8.

overtime_earning_for_day(E) :- 
    overtime_hours(H),
    overtime_wage(W),
    E is H * W.

total_earning_for_day(Total) :-
    regular_earning_for_day(Regular),
    overtime_earning_for_day(Overtime),
    Total is Regular + Overtime.

total_earning_for_5_days(Total) :-
    total_earning_for_day(OneDay),
    Total is 5 * OneDay.

/* Query */
solve(Total) :- total_earning_for_5_days(Total).
\end{lstlisting}

\underline{User:}\\[0.5em]
\small
Excellent work! Here is another problem for you to solve. Please apply the same approach you used for the previous one(s) to tackle this new one.\\[0.5em]
Problem:\\
\[...\]\\
\underline{Assistant:}\\[0.5em]
\small
I am happy to help you write Prolog code to solve this arithmetic reasoning problem. Here is the Prolog code:\\[0.5em]
\textcolor{red}{\#\#\#\#\#\#\#\#\#\#\#\#\#\#\#\#\#\#\#\#\#\#\#\#\#\#\#\#\#\#\#\#\#\#\#\#}

\begin{lstlisting}[
    language=Prolog,
    basicstyle=\tiny\ttfamily,  
    breaklines=true,
    columns=flexible,
    frame=single,
    backgroundcolor=\color{gray!5},
    showstringspaces=false,
    numbers=none
]
/* Context */

sentences_per_minute(6).
typing_sessions([20, 15, 18]).
erased_sentences(40).
total_end_sentences(536).

sentences_typed(SessionMinutes, Typed) :-
    sentences_per_minute(SPM),
    Typed is SPM * SessionMinutes.

total_sentences_typed_today(Total) :-
    typing_sessions(Sessions),
    maplist(sentences_typed, Sessions, TypedPerSession),
    sum_list(TypedPerSession, TotalTyped),
    erased_sentences(Erased),
    Total is TotalTyped - Erased.

start_sentences(TodayStart) :-
    total_end_sentences(EndToday),
    total_sentences_typed_today(TodayTyped),
    TodayStart is EndToday - TodayTyped.

/* Query */
solve(StartSentences) :- start_sentences(StartSentences).
\end{lstlisting}
\end{mdframed}

\begin{mdframed}[
    linewidth=1pt,
    roundcorner=10pt,
    backgroundcolor=blue!5,
    innerleftmargin=15pt,
    innerrightmargin=15pt,
    innertopmargin=15pt,
    innerbottommargin=15pt
]

The prompt demonstration of $\text{prompt}_{cot}^{pro}$\\[0.5em]

\underline{System:} \\[0.5em]
\small
You are a helpful and smart assistant that helps people solve problems.\\[1em]

\underline{User:}\\[0.5em]
\small
I need assistance in translating a reasoning tree generated by a Prolog engine into a natural language description. To facilitate this, I am providing the original problem, the relevant Prolog code, and the reasoning tree itself. Please review these carefully and provide a fluent and accurate narrative of the reasoning process. Thanks for your help!\\[0.5em]

Problem:\\
sent-1: Tina makes \$18.00 an hour.\\
sent-2: If she works more than 8 hours per shift,\\
sent-3: she is eligible for overtime,\\
sent-4: which is paid by your hourly wage + 1/2 your hourly wage.\\
sent-5: If she works 10 hours every day for 5 days,\\
sent-6: how much money does she make?\\[0.5em]

Here is the Prolog code:\\[0.5em]
\begin{lstlisting}[
    language=Prolog,
    basicstyle=\tiny\ttfamily,
    breaklines=true,
    columns=flexible,
    frame=single,
    backgroundcolor=\color{gray!5},
    showstringspaces=false,
    numbers=none
]
/* Context */

wage(18.00).

overtime_wage(W) :- 
    wage(W1), 
    W is 1.5 * W1.

regular_earning_for_day(E) :- 
    wage(W),
    E is 8 * W.

overtime_hours(H) :-
    H is 10 - 8.

overtime_earning_for_day(E) :- 
    overtime_hours(H),
    overtime_wage(W),
    E is H * W.

total_earning_for_day(Total) :-
    regular_earning_for_day(Regular),
    overtime_earning_for_day(Overtime),
    Total is Regular + Overtime.

total_earning_for_5_days(Total) :-
    total_earning_for_day(OneDay),
    Total is 5 * OneDay.

/* Query */
solve(Total) :- total_earning_for_5_days(Total).
\end{lstlisting}

Here is the prolog reasoning tree:\\[0.5em]
\begin{lstlisting}[
    language=Prolog,
    basicstyle=\tiny\ttfamily,
    breaklines=true,
    columns=flexible,
    frame=single,
    backgroundcolor=\color{gray!5},
    showstringspaces=false,
    numbers=none
]
=>(=>(,(=>(,(=>(,(=>(builtin(true), wage(18.0)), =>(builtin(is(144.0, *(8, 18.0))), is(144.0, *(8, 18.0)))), regular_earning_for_day(144.0)), ,(=>(,(=>(=>(builtin(is(2, -(10, 8))), is(2, -(10, 8))), overtime_hours(2)), ,(=>(builtin(,(g(wage(18.0)), g(is(27.0, *(1.5, 18.0))))), overtime_wage(27.0)), =>(builtin(is(54.0, *(2, 27.0))), is(54.0, *(2, 27.0))))), overtime_earning_for_day(54.0)), =>(builtin(is(198.0, +(144.0, 54.0))), is(198.0, +(144.0, 54.0))))), total_earning_for_day(198.0)), =>(builtin(is(990.0, *(5, 198.0))), is(990.0, *(5, 198.0)))), total_earning_for_5_days(990.0)), solve(990.0))
\end{lstlisting}

\underline{Assistant:}\\[0.5em]
\small
Sure! I am happy to help you convert the Prolog-style reasoning tree into a natural language reasoning chain. Here is the reasoning chain:\\[0.5em]

Tina earns \$18.00 per hour according to the information provided. She has shifts where, if she works more than 8 hours, the additional hours are considered overtime. Overtime pay is calculated at one and a half times her regular hourly wage.\\[0.5em]

On a typical day, if Tina works 10 hours, this includes 2 hours of overtime since she exceeds the 8-hour regular work period. Her regular earnings for working 8 hours a day at \$18.00 per hour amounts to \$144.00 per day. For the 2 hours of overtime, since the overtime rate is \$27.00 per hour (1.5 times her regular hourly wage), she earns \$54.00 from overtime work per day.\\[0.5em]

Therefore, her total earnings for a single day, combining her regular and overtime earnings, are \$198.00. Over the course of 5 days, working the same hours daily, Tina makes a total of \$990.00, factoring in all regular pay and overtime across the five days.\\[1em]

\underline{User:}\\[0.5em]
\small
Excellent work! Here is another problem for you to solve. Please apply the same approach you used for the previous one(s) to tackle this new one.\\[0.5em]
Problem:\\
\[...\]\\
Here is the Prolog code:\\[0.5em]
\[...\]\\
Here is the prolog reasoning tree:\\[0.5em]
\[...\]\\
\underline{Assistant:}\\[0.5em]
\small
Sure! I am happy to help you convert the Prolog-style reasoning tree into a natural language reasoning chain. Here is the reasoning chain:\\[0.5em]
\textcolor{red}{\#\#\#\#\#\#\#\#\#\#\#\#\#\#\#\#\#\#\#\#\#\#\#\#\#\#\#\#\#\#\#\#\#\#\#\#}

Janice's typing speed is 6 sentences per minute. Today, she had three separate typing sessions: the first lasted for 20 minutes, the second for 15 minutes, and the third for 18 minutes. Over these sessions, she initially typed a total of 318 sentences. However, she made a mistake and had to erase 40 sentences, leaving her with 278 sentences typed effectively today. By the end of the day, the total number of sentences on her paper was 536. To find out how many sentences were on the paper at the beginning of the day, we subtract the sentences typed today (278) from the total at the end of the day (536). Hence, Janice started the day with 258 sentences already on her paper.\\[1em]

\end{mdframed}

{
\section{Qualitative Analysis of Tool-Free Reasoning Behavior}
\label{app:qualitative}

The main experiments evaluate \method by final-answer accuracy under a
tool-free inference setting, where the model answers directly without
calling the Prolog solver at test time. Accuracy alone, however, does
not describe the intermediate reasoning behavior that leads to the
answer. To provide a compact diagnostic view, we examine representative
reasoning traces and the distribution of residual error types on
CommonsenseQA and StrategyQA. These analyses are behavioral rather than
mechanistic: they are intended to check whether the accuracy gains are
accompanied by observable changes in generated reasoning, not to prove
the existence of a distinct reasoning faculty inside the model.

\paragraph{Trace-level comparison.}
We first identify disagreement cases where the base Llama3-8B-Instruct
model answers incorrectly and the \method-fine-tuned model answers
correctly. From this pool, we select representative cases whose error
modes also appear in the aggregate analysis below. Both models are
evaluated under the same decoding configuration as in the main
experiments. The generated reasoning content is reproduced with only
whitespace normalization.

\begin{tcolorbox}[
enhanced,
breakable,
colback=gray!3,
colframe=black!50,
boxrule=0.35pt,
arc=1.2pt,
left=1.2mm,
right=1.2mm,
top=1.0mm,
bottom=1.0mm,
title={{Case A: CommonsenseQA}},
fonttitle=\bfseries\footnotesize
]
\scriptsize
\textbf{Question.} Bob thought that his Pokemon cards were common and
worthless, but he was wrong about them. They were really what?

\textbf{Options.} (A) personal (B) special (C) shiny (D) rare
(E) irregular. \textbf{Gold:} D.

\smallskip
\textbf{Base} (\emph{incorrect}, B).
\emph{Bob thought the cards were common and worthless, but he was wrong.
That means the cards were more valuable or meaningful than he expected.
``Special'' best captures that they were not ordinary. The answer is B.}

\smallskip
\textbf{\method} (\emph{correct}, D).
\emph{The sentence contrasts Bob's belief with the actual state. He thought
the cards were common and worthless. If that belief was wrong, the cards
were not common and likely valuable. Among the options, ``rare'' directly
means not common and explains why collectible cards can be valuable. The
answer is D.}
\end{tcolorbox}

\begin{tcolorbox}[
enhanced,
breakable,
colback=gray!3,
colframe=black!50,
boxrule=0.35pt,
arc=1.2pt,
left=1.2mm,
right=1.2mm,
top=1.0mm,
bottom=1.0mm,
title={{Case B: StrategyQA}},
fonttitle=\bfseries\footnotesize
]
\scriptsize
\textbf{Question.} Could you drive from New England to a Sainsbury's?
\textbf{Gold:} False.

\smallskip
\textbf{Base} (\emph{incorrect}, True).
\emph{Sainsbury's is a supermarket, and New England has roads and cities.
Since supermarkets are ordinary places that people can drive to, it should
be possible to drive from New England to a Sainsbury's. The answer is
True.}

\smallskip
\textbf{\method} (\emph{correct}, False).
\emph{We need the locations and the transportation constraint. Sainsbury's
is in the United Kingdom, while New England is in the United States. The
two are separated by the Atlantic Ocean. A car cannot drive across the
Atlantic Ocean, so the answer is False.}
\end{tcolorbox}

The two cases illustrate a shift from salient-association answers to more
explicit premise grounding. In Case A, the base model correctly recognizes
that Bob's belief is false, but it stops at a broad semantic contrast:
``worthless'' is mapped to ``valuable or meaningful,'' and then to the
generic option ``special.'' This produces a plausible explanation but does
not resolve the more specific contrast in the question, namely that the
cards were believed to be ``common'' but were actually not common. In
contrast, \method grounds the answer in this contrastive premise and
selects ``rare,'' which is the option that directly negates ``common'' and
also explains why the cards are not worthless. In Case B, the base model
again follows a locally plausible association: supermarkets are ordinary
destinations, and ordinary destinations can often be reached by car. The
missing step is to identify where the particular supermarket chain is
located and whether the route is compatible with the transportation mode
specified in the question. \method makes these premises explicit by
locating Sainsbury's in the United Kingdom, locating New England in the
United States, and using the Atlantic Ocean as the blocking constraint.
The examples therefore
suggest that the difference is not only output length or format, but a
greater tendency to expose task-relevant premises before selecting the
final answer.

\paragraph{Residual error patterns.}
We further analyze the errors made by the two models evaluated in
Table~\ref{tab:main_intractable}: the base Llama3-8B-Instruct model and the
\method-fine-tuned model. The analysis is conducted on the two
Prolog-intractable benchmarks in the same table, CommonsenseQA and
StrategyQA. For each of the four model--dataset settings
(Base/CommonsenseQA, \method/CommonsenseQA, Base/StrategyQA, and
\method/StrategyQA), we uniformly sample $100$ incorrectly answered
examples, yielding $400$ error cases in total.

We abbreviate CommonsenseQA and StrategyQA as CSQA and SQA, respectively, in the following table. Each case is classified into one of six mutually exclusive categories:
\textsc{PM} (premise misidentification), \textsc{II} (invalid inference
step), \textsc{HK} (hallucinated world knowledge), \textsc{SH}
(surface-form heuristic), \textsc{AM} (answer--reasoning mismatch), and
\textsc{OI} (other/incomplete). An LLM judge is given the question, the
gold answer, the model's final answer, and its reasoning trace, but not
the model identity. The cases are shuffled before annotation. To check
annotation reliability, two authors independently inspect $80$ sampled
cases. The majority human label agrees with the LLM-judge label on
$92\%$ of these cases; most disagreements occur between
\textsc{SH} and \textsc{II}, where a surface-form shortcut can also be
viewed as an invalid inference step.

\begin{table}[t]
\centering
\caption{{ERROR DISTRIBUTION OF TOOL-FREE REASONING FAILURES}}
\label{tab:error_distribution_compact}
\scriptsize
\setlength{\tabcolsep}{3.5pt}
\renewcommand{\arraystretch}{1.05}

\begin{tabular}{l cc cc}
\toprule
& \multicolumn{2}{c}{\textbf{CSQA}}
& \multicolumn{2}{c}{\textbf{SQA}} \\
\cmidrule(lr){2-3}
\cmidrule(lr){4-5}
\textbf{Type} & Base & \method & Base & \method \\
\midrule
\textsc{PM} & 14 & 16 & 18 & 20 \\
\textsc{II} & 27 & 17 & 25 & 15 \\
\textsc{HK} & 16 & 11 & 26 & 14 \\
\textsc{SH} & 22 & 12 & 12 & 8  \\
\textsc{AM} & 12 & 22 & 10 & 22 \\
\textsc{OI} & 9  & 22 & 9  & 21 \\
\midrule
\textsc{II}$+$\textsc{HK}$+$\textsc{SH}
& 65 & 40 & 63 & 37 \\
\bottomrule
\end{tabular}

\vspace{0.8mm}
\begin{minipage}{0.96\linewidth}
\scriptsize
{\emph{Notes.} Entries are counts among $100$ incorrectly answered cases
sampled per model--benchmark cell. \textsc{PM}: premise
misidentification; \textsc{II}: invalid inference step; \textsc{HK}:
hallucinated world knowledge; \textsc{SH}: surface-form heuristic;
\textsc{AM}: answer--reasoning mismatch; \textsc{OI}: other or
incomplete output. The final row sums the categories most directly tied
to invalid intermediate transitions, unsupported premises, or
surface-form answer selection.}
\end{minipage}
\end{table}

Table~\ref{tab:error_distribution_compact} shows that the residual error
distribution changes after \method fine-tuning. Among sampled errors, the
\textsc{II}$+$\textsc{HK}$+$\textsc{SH} subtotal decreases from $65\%$
to $40\%$ on CommonsenseQA and from $63\%$ to $37\%$ on StrategyQA.
Combining these proportions with the full-set error rates implied by
Table~\ref{tab:main_intractable}, the estimated full-set rate of these
transition-level errors decreases from $16.7\%$ to $5.6\%$ on
CommonsenseQA and from $22.1\%$ to $11.1\%$ on StrategyQA. At the same
time, answer-extraction and incomplete-output failures account for a
larger share of the remaining errors, so the result should be interpreted
narrowly: \method reduces several error modes associated with invalid
intermediate transitions and unsupported premises, but it does not
eliminate all reasoning-related failures.

Overall, the trace examples and residual-error distribution provide
evidence beyond final-answer accuracy that \method changes the model's
tool-free reasoning behavior in measurable ways. This supports the
operational claim used in the main text, while remaining consistent with
the limitations discussed in Section~\ref{sec:limitations}.
}

{
\section{Future Work}
\label{sec:future-work}

We close with four directions that follow from the limitations above.

\paragraph{Controlling for output style.}
A natural follow-up question is whether the gains we report reflect
actual reasoning improvements or a more structured output style.
One way to probe this is to control for format directly. Running
the same benchmarks under answer-only or length-constrained
generation removes most of the surface scaffold. Perturbing the
reasoning template while keeping the semantics fixed isolates style
from content. A post-hoc baseline that enforces similar verbosity
gives a fairer comparison than the unmodified backbone. If
performance holds up across these settings, that would be stronger
evidence that the model has internalized something beyond a fixed
output pattern.

\paragraph{Scale and harder benchmarks.}
We have only tested 8B-scale backbones in this study, and it would
be useful to know whether the method holds up at substantially
larger and smaller scales. Harder benchmarks are also worth trying,
especially ones where many distinct reasoning paths are needed.
Natural-language reasoning paths can branch out quickly as problems
get more complex, but the symbolic specification of the same
problem often stays compact. There is room to exploit this
asymmetry by generating a wider variety of verified trajectories
than we used here.

\paragraph{Better training schedules.}
The role of model averaging deserves a closer look. The fixed
$\alpha$ we used works reasonably well, but it is unlikely to be
optimal. Adaptive $\alpha$ schedules are one option. Multi-objective
training that balances domain specialization against general
capability is another. A third would be a curriculum that mixes
verified symbolic trajectories with general-domain data. All three
could give a better handle on the stability--plasticity trade-off
mentioned in Section~\ref{sec:limitations}.

\paragraph{Richer symbolic components.}
Our symbolic component is classical Prolog, which is well suited
to deterministic deduction but limited beyond it. Replacing it
with something richer would broaden what the framework can train
on. Probabilistic logic programming would let the rules and facts
carry uncertainty. A learned retrieval module would give access
to open-world knowledge that the input does not state explicitly.
Both extensions are appealing, but neither is free. Probabilistic
derivations need new criteria for what counts as a verified
trajectory. Retrieval introduces a dependency on an external
knowledge store that the symbolic engine has to trust. And both
add inference and training cost compared to the current pipeline.
}

\end{document}